\documentclass[11pt]{article}
\usepackage[margin=1in]{geometry}
\usepackage{times}
\usepackage{amsmath,amssymb,amsthm}
\usepackage[round]{natbib}          
\usepackage{graphicx,url}

\newtheorem{theorem}{Theorem}
\newtheorem{lemma}[theorem]{Lemma}
\newtheorem{proposition}[theorem]{Proposition}

\theoremstyle{definition}
\newtheorem{definition}[theorem]{Definition}

\title{How Much Distribution Shift Can a Clustering Withstand?\\
Exact Split and Merge Certificates for Structured Transports}

\author{%
Alisina Bayati \qquad Srinivasa M. Salapaka\\
{\small Department of Mechanical Science and Engineering and Coordinated Science Laboratory}\\
{\small University of Illinois Urbana-Champaign, Urbana, IL, USA}\\
{\small \texttt{\{abayati2,salapaka\}@illinois.edu}}
\and
Shahriar Talebi\\
{\small Department of Mechanical and Aerospace Engineering}\\
{\small University of California, Los Angeles, Los Angeles, CA, USA}\\
{\small \texttt{s.talebi@ucla.edu}}}
\date{}

\newcommand{\R}{\mathbb{R}}

\newcommand{\U}{\mathcal{U}}

\usepackage{xcolor}
\usepackage{enumitem}
\usepackage[colorlinks,allcolors=blue]{hyperref}  
\newif\ifshowtodos \showtodosfalse   

\begin{document}

\maketitle

\begin{abstract}%
In clustering, the distribution of the population being clustered is often uncertain or changes over time, so the true distribution may shift away from the nominal distribution used to form the clustering. Once the clustering is committed and each member is assigned to a cluster, such shifts may change the distribution of each cluster. They can make a cluster too spread out for one representative at a given resolution, motivating a split, or bring distinct clusters close enough to share a representative at that resolution, motivating a merge. We make these split and merge events precise using a local stability criterion built on the phase-transition analysis in deterministic annealing. This notion of stability enables us to quantify how robust a clustering is against such events. Furthermore, motivated by applications in which uncertainty and population change are structured, we study shifts whose displacements are restricted to a prescribed subspace. The smallest structured shifts that reach the respective thresholds define the split and merge radii, for which we obtain exact semidefinite programming (SDP) reformulations. The split SDP has only two scalar decision variables. These convex reformulations enable us to combine the radii into a robustness certificate for the full clustering that gives the size of the smallest structured shift that reaches any monitored split or merge threshold.
\end{abstract}


\section{Introduction}
\label{sec:introduction}

Clustering addresses a fundamental problem: how to represent or serve a
population effectively with a limited number of representatives. This problem arises in applications ranging from data compression and learning to sensor coverage and resource allocation \citep{rose_deterministic_1998,mavridis_annealing_2023,mavridis_multi-resolution_2023,cortes_coverage_2004,sharma_scalable_2008,bayati_control_2025}. A clustering partitions points in \(\R^d\) into groups, each summarized by a representative or assigned a shared model, controller, or resource. The quality of this representation is commonly measured through
\emph{distortion}, typically defined as the average squared Euclidean
distance between members and their representatives.

Once a clustering is committed, uncertainty in the original data (e.g.,
measurement errors) or later changes in the population may cause its
distribution to differ from the \emph{nominal distribution} used to form
the clustering. We call this difference a \emph{distribution shift}. Such
a shift may change the number of representatives the population needs. A
cluster may become too dispersed for one representative, motivating a
split, or several clusters may come close enough to share one, motivating a merge. These revisions can incur both operational and computational costs: splitting may require additional resources, merging may require reconfiguring assignments or shared models, and finding a new globally optimal clustering is generally NP-hard \citep{aloise_np-hardness_2009}. It is therefore valuable to ask in advance: \emph{How much distribution shift can a labeled clustering withstand before a split or a merge is warranted?}

To make this question precise, we must account for resolution: the number
of representatives a population needs depends on the resolution at which
it is viewed. At a coarse resolution, one representative may suffice for
the entire population, whereas at a fine enough resolution, each distinct
member of a finite population may need its own. Between these extremes,
critical resolutions mark where a cluster splits into finer clusters or
where several clusters merge into one. One way to formalize this notion
of resolution is through deterministic annealing (DA), which uses a
temperature parameter \(T\) to balance distortion against the entropy of
cluster assignments \citep{rose_deterministic_1998,srivastava_persistence_2019}. The
resulting objective is called the \emph{free energy}. At sufficiently
high temperatures, coincident representatives at the population mean
form a local minimum of the free energy, resulting in a single effective
cluster. As \(T\) decreases, the resolution increases, and the coincident
configuration undergoes a bifurcation at a \emph{critical temperature}.
Just below this temperature, the configuration becomes a saddle point, and
separating the representatives along certain directions lowers the free
energy, allowing distinct clusters to emerge and reveal finer structure in the population \citep{rose_deterministic_1998}. We refer to this local behavior as the \emph{stability} of a distribution at temperature \(T\) (Definition~\ref{def:da-stability}), which we apply to individual clusters and to collections of clusters.

To assess a given clustering without changing its assignments, we keep
cluster labels attached to population members as the distribution
shifts, preserving each cluster's mass. At a fixed temperature \(T\), a
distribution reaches the \emph{stability threshold} when \(T\) is its
critical temperature. A \emph{split event} occurs when a shift brings a
cluster's initially stable distribution to this threshold, and a
\emph{merge event} occurs when a shift brings a specified collection's
initially unstable pooled distribution to it. The events to monitor, for
example splits of selected clusters or only pairwise merges, are
prescribed by the application. For squared Euclidean distortion, Rose's
phase-transition analysis \citep{rose_deterministic_1998} gives the
second-order stability criterion \(\lambda_{\max}(C(q))<T/2\), where
\(C(q)\) is the covariance of the distribution \(q\), with equality
defining the stability threshold. This criterion involves only the
covariance of the distribution being tested, so it can be used to assess
a clustering regardless of how that clustering was obtained.

\begin{figure}[t]
\centering
\includegraphics[width=\textwidth]{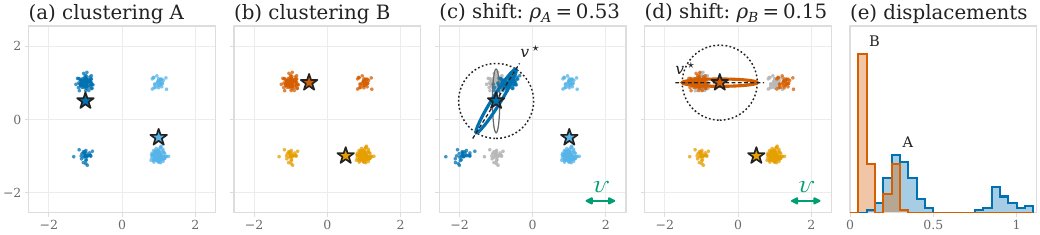}
\caption{Two clusterings of the same population with identical distortion need very different horizontal shifts to reach the split threshold. 
(a),~(b)~Candidate clusterings A and B; stars mark the cluster centroids.
(c),~(d)~Minimum-cost horizontal shifts that bring one cluster in each
clustering to the split threshold at the
same \(T\); grey shows the nominal cluster and
\(v^\star\) the split direction. The threshold is reached when the
covariance ellipse, with longest semiaxis
\(\sqrt{\lambda_{\max}(C(q))}\), touches the dotted circle of radius
\(\sqrt{T/2}\). (e)~Histograms of point displacement magnitudes under the shifts
in (c) and (d); each split radius is the corresponding
root-mean-square displacement.}
\label{fig:intro}
\end{figure}

Quantifying the distance to a stability threshold requires specifying
which population shifts are allowed and how their magnitude is measured. We measure a distribution shift by the square root of the smallest expected squared Euclidean displacement needed to transport the nominal distribution to the shifted distribution. Without restrictions on the displacements, this is the \(2\)-Wasserstein distance \citep{peyre_computational_2019}.

In many applications, shifts are \emph{structured}. For instance, disturbances and attacks in cyber-physical systems act through specific actuator or sensor channels \citep{fawzi_secure_2014}, changes in feature vectors may affect only selected attributes while others remain fixed \citep{ustun_actionable_2019}, and adversarial image perturbations may be restricted to a low-frequency subspace \citep{guo_low_2020}. We model this structure by requiring displacements to lie in a prescribed linear subspace \(\U\subseteq\R^d\), with \(\U=\R^d\) recovering unrestricted shifts. The \emph{split radius} and \emph{merge radius} are the smallest structured shifts, measured by this transport distance, that reach the respective events. These radii quantify the robustness of individual clusters or collections to specific events. To assess the clustering as a whole, we combine them into an overall robustness measure. Here, we adopt a worst-case approach: we define a transport distance for the entire
labeled population and use the distance to the nearest monitored event
as a robustness \emph{certificate}. Every smaller structured
shift is guaranteed to avoid all monitored events.

Such a certificate can also help compare candidate clusterings
before deployment. Figure~\ref{fig:intro} compares two clusterings of the
same population with identical nominal distortion. Under horizontal
shifts at the same temperature, A's split radius is more than three
times B's, revealing a substantial
difference in robustness that distortion alone cannot capture.
Comparing certificates across choices of \(\U\) shows how a
clustering's robustness depends on the allowed displacement directions.

Structured shifts introduce a geometric difficulty:
\emph{restricting displacements to \(\U\) does not restrict the direction
of instability to \(\U\)} (notice how \(v^\star\) is not aligned with
\(\U\) in Figure~\ref{fig:intro}(c)). Displacements within \(\U\) can change the variance along directions with components in both \(\U\) and
\(\U^\perp\), so the structured problem cannot generally be solved by
projecting the unrestricted problem onto \(\U\). However, these
displacements leave each point's component in \(\U^\perp\) unchanged.
Thus, a collection whose variance along some direction in
\(\U^\perp\) already exceeds the threshold \(T/2\) cannot reach the
merge threshold at any transport budget. An initially stable cluster,
on the other hand, can always reach the split threshold when
\(\U\ne\{0\}\).

For fixed \(T\) and \(\U\), we show that each event radius depends only on the covariance of the tested cluster or pooled collection. Our contributions are fourfold:
\begin{enumerate}
\item \textbf{Exact moment reduction.}
We reduce the split and merge radius problems from optimization over infinite-dimensional structured transport plans to finite-dimensional optimization over the displacement covariance and the cross-covariance between nominal points and their displacements. The reduction is exact: every feasible pair of moments is realized by a structured transport plan. The results hold for arbitrary nominal distributions with finite second moments.

\item \textbf{Split radius.}
For each feasible direction, we derive the minimum structured transport cost needed to reach the split threshold along it. Although minimizing this cost over directions is nonconvex, we compute the squared split radius through an exact SDP reformulation with two scalar decision variables. We recover an optimal split direction and construct an attaining structured transport plan.

\item \textbf{Merge radius.}
For a specified collection of clusters, we formulate the squared merge radius as an exact SDP. We show that a merge is feasible if and only if the nominal pooled distribution has variance at most \(T/2\) along every direction in \(\U^\perp\). Whenever this condition holds, a deterministic affine structured transport map attains the minimum cost.

\item \textbf{Clustering certificate.}
We define a label-preserving structured transport distance on clusterings and show that the distance to the nearest monitored event equals the minimum of the event radii, each scaled by the square root of the affected population mass. If this certificate is finite, an attaining distribution shift exists which can cause at least one of the monitored events to happen, provided its corresponding transport budget.
\end{enumerate}


\paragraph{Related work.}
The interpretation of DA temperature as resolution is also used by \citet{srivastava_persistence_2019} to assess clusterings through the range of resolutions over which they persist, and by online DA methods to progressively increase model complexity \citep{mavridis_annealing_2023,mavridis_multi-resolution_2023}. While these works vary the resolution, we ask a complementary question:
with the clustering and resolution fixed, what is the smallest
distribution shift that reaches a split or merge threshold?

Resampling stability asks whether rerunning an algorithm on a new sample produces the same partition \citep{ben-david_sober_2006,von_luxburg_clustering_2010}, and perturbation stability assumes that an optimal clustering must remain unchanged under bounded multiplicative perturbations of pairwise distances \citep{awasthi_center-based_2012}. Our certificate concerns prescribed split and merge events for a given labeled clustering, without requiring optimality or rerunning the algorithm.

Wasserstein distributionally robust optimization (DRO) chooses decisions that minimize worst-case expected cost over a Wasserstein ball \citep{mohajerin_esfahani_data-driven_2018,gao_distributionally_2023}. For \(k\)-means clustering, \citet{malik_distributionally_2026} use this framework to design representatives that minimize worst-case distortion under a prescribed shift budget. We instead assess a given clustering and quantify its robustness, rather than designing representatives.

Our use of displacement covariances and cross-covariances between
nominal points and their displacements is related to moment formulations
in covariance steering \citep{chen_optimal_2016,balci_exact_2022}. Our setting differs from transport between spaces of unequal dimensions \citep{mccann_optimal_2020}, as both nominal and shifted distributions lie in \(\R^d\), while their displacements are restricted to a subspace \(\U\). Lastly, our transport plans preserve cluster labels and each point's component in \(\U^\perp\). The resulting distance is therefore a conditional Wasserstein distance in the sense of \citet{chemseddine_conditional_2025}, conditioned on the label and the unchanged orthogonal component.

\paragraph{Notation.}
Vector norms are Euclidean, \(I_d\) is the identity matrix, and
\(\mathbb{S}^d\) denotes the real symmetric \(d\times d\) matrices.
For \(A,B\in\mathbb{S}^d\), \(\lambda_{\max}(A)\) is the largest
eigenvalue of \(A\), and \(A\preceq B\) means that \(B-A\) is
positive semidefinite. We denote the Moore--Penrose pseudoinverse
by \(A^\dagger\). For a set \(E\), \(2^E\) denotes its power set,
the collection of all subsets of \(E\).
The set \(\mathcal{P}_2(\R^d)\) consists of probability distributions with finite second moments; \(\mu(q)\) and \(C(q)\) denote the mean and covariance of \(q\). For \(q_0,q\in\mathcal{P}_2(\R^d)\), \(\Gamma(q_0,q)\) denotes the set of couplings, or transport plans, of \(q_0\) and \(q\), i.e., probability distributions on \(\R^d\times\R^d\) with first marginal \(q_0\) and second marginal \(q\). The squared \(2\)-Wasserstein distance is
\begin{equation}
\label{eq:wasserstein-distance}
W_2^2(q_0,q)
:=
\inf_{\gamma\in\Gamma(q_0,q)}
\int\|z-x\|^2\,\gamma(dx,dz).
\end{equation}

\section{Background and Problem Formulation}
\label{sec:preliminaries}
\paragraph{Clustering and free energy.}
Squared-error clustering chooses representatives
\(Y=(y_1,\ldots,y_k)\) to minimize the distortion
\(\mathcal{D}(Y;q):=\int\min_j\|x-y_j\|^2\,q(dx)\), corresponding to assigning each point to one (the nearest) representative. Deterministic annealing (DA) replaces these binary assignments with probabilities and subtracts \(T>0\) times their entropy. Minimizing the resulting \emph{free energy} \citep{rose_deterministic_1998} over the assignment probabilities gives
\begin{equation}
\label{eq:free-energy}
\mathsf{F}_T(Y;q)
:=
-T\int\log\Bigl(
\sum_{j=1}^k e^{-\|x-y_j\|^2/T}
\Bigr)\,q(dx).
\end{equation}
As \(T\downarrow0\), \(\mathsf{F}_T(Y;q)\) converges to \(\mathcal{D}(Y;q)\), uniformly in \(Y\). (see Appendix~\ref{app:da-free energy}).


Following the DA formulation, we fix a temperature \(T>0\) to specify the
resolution and use the free energy to decide when a single representative
at the mean is locally optimal for a distribution. Concretely, for
\(q\in\mathcal{P}_2(\R^d)\), we place \(k\ge2\) representatives at
\(\mu(q)\) and ask whether slightly separating them lowers the free
energy \eqref{eq:free-energy}. This coincident configuration
\(Y_\mu(q):=(\mu(q),\ldots,\mu(q))\) is always first-order stationary, i.e., \(\nabla_Y\mathsf{F}_T(Y_\mu(q);q)=0\) (Appendix~\ref{app:da-covariance-stability}), so we examine its Hessian.

\begin{definition}[Stability]
\label{def:da-stability}
At temperature \(T>0\) and for any \(k\ge2\), a distribution
\(q\in\mathcal{P}_2(\R^d)\) is \emph{stable} if
\(\nabla_Y^2\mathsf{F}_T(Y_\mu(q);q)\succ0\), \emph{unstable} if this
Hessian has a negative eigenvalue, and \emph{critical} if it is positive
semidefinite but singular (we call it critical because \(T\) is then its
critical temperature).
\end{definition}
This classification does not depend on \(k\) (see Lemma~\ref{lem:da-covariance}), and critical distributions are exactly those at the stability threshold.

A \emph{split} event occurs when an initially stable cluster distribution
reaches the stability threshold, and a \emph{merge} event occurs when the
initially unstable pooled distribution of a collection reaches it. Both
events include critical distributions, since an arbitrarily small
(unrestricted) perturbation of a critical \(q\) in the \(2\)-Wasserstein sense can make it stable (e.g., a contraction about its mean) or unstable (e.g., an expansion). The covariance criterion in Lemma~\ref{lem:da-covariance} makes this clear.

To determine which shifts can reach these thresholds and at what cost, we
first define structured transport plans. Let \(\U\subseteq\R^d\) be a
prescribed linear subspace with \(\U\neq\{0\}\). A transport plan is
\emph{\(\U\)-structured} if its displacements lie in \(\U\) almost
surely. For \(q_0,q\in\mathcal{P}_2(\R^d)\), define
\begin{equation}
\label{eq:Gamma_U}
\Gamma_{\U}(q_0,q)
:=
\bigl\{\gamma\in\Gamma(q_0,q):
z-x\in\U\ \text{for }\gamma\text{-a.e.\ }(x,z)\bigr\},
\end{equation}
and the \emph{structured transport distance}
\begin{equation}
\label{eq:restricted-wasserstein}
W_{2,\U}^2(q_0,q)
:=
\inf_{\gamma\in\Gamma_{\U}(q_0,q)}
\int\|z-x\|^2\,\gamma(dx,dz),
\end{equation}
the least cost of transporting \(q_0\) to \(q\) with \(\U\)-structured
plans. We use the convention \(\inf\varnothing=+\infty\); for
\(\U=\R^d\), \(W_{2,\U}=W_2\). The distance is finite if and only if
\(q_0\) and \(q\) have the same marginal on \(\U^\perp\), and if so, an
optimal plan exists. When the \(\U^\perp\) marginals agree, \(W_{2,\U}\) is the
conditional Wasserstein distance of \citet{chemseddine_conditional_2025},
with the \(\U^\perp\)-component as the conditioning variable.
Appendix~\ref{app:conditional-wasserstein} proves both facts
(Lemma~\ref{lem:finiteness}) and makes this connection precise.

\paragraph{The Split and Merge Problems.}
For a nominal cluster distribution \(q_0\in\mathcal{P}_2(\R^d)\), the
\emph{split radius} is
\begin{equation}
\label{eq:split-radius}
\rho_{\mathrm{s},\U}(q_0;T)
:=
\inf_{q\in\mathcal{P}_2(\R^d)}
\Bigl\{
W_{2,\U}(q_0,q):
\nabla_Y^2\mathsf{F}_T(Y_\mu(q);q)\not\succ0
\Bigr\},
\end{equation}
which is positive when \(q_0\) is stable. Similarly, for the nominal
pooled distribution \(q_0\in\mathcal{P}_2(\R^d)\) of a collection of
clusters, the \emph{merge radius} is
\begin{equation}
\label{eq:merge-radius}
\rho_{\mathrm{m},\U}(q_0;T)
:=
\inf_{q\in\mathcal{P}_2(\R^d)}
\Bigl\{
W_{2,\U}(q_0,q):
\nabla_Y^2\mathsf{F}_T(Y_\mu(q);q)\succeq0
\Bigr\},
\end{equation}
which is positive when \(q_0\) is unstable. Both radii equal zero when \(q_0\) is critical.

The radii above quantify the robustness of a single cluster to
splitting or of a collection of clusters to merging into one. To assess the whole clustering against a set of monitored events, we need a distance on population shifts that accounts for each cluster's displacement and mass.

Consider a labeled clustering with \(M\) clusters, indexed by
\([M]:=\{1,\ldots,M\}\). Let \(m_j>0\) denote the mass of cluster
\(j\), i.e., its fraction of the population, with
\(\sum_j m_j=1\), and let \(q_0^j,q^j\in\mathcal{P}_2(\R^d)\)
denote the nominal and shifted distributions conditional on the
cluster label \(j\). Write
\(\mathbf{q}_0:=(q_0^1,\ldots,q_0^M)\) and
\(\mathbf{q}:=(q^1,\ldots,q^M)\) for the corresponding tuples of
cluster distributions.  Since labels stay attached to the transported mass, each \(m_j\) is preserved, and each cluster moves by its own \(\U\)-structured plan. The population transport cost is the mass-weighted sum of the cluster transport costs. For fixed shifted distributions, the plans can be optimized independently, so the minimum population cost is the mass-weighted sum of the minimum cluster costs. We therefore define the \emph{label-preserving}
\(\U\)-structured transport distance
\begin{equation}
\label{eq:clustering-distance}
\mathsf{W}_\U^2(\mathbf{q}_0,\mathbf{q})
:=\sum_{j=1}^M m_j\,W_{2,\U}^2(q_0^j,q^j).
\end{equation}
When finite, it is a conditional \(2\)-Wasserstein distance, conditioned
on both the cluster label and the unchanged \(\U^\perp\)-component.

Let \(\mathcal{S}\subseteq[M]\) index the clusters
monitored for splitting, and let \(\mathcal{A}\subseteq2^{[M]}\) be the
family of collections monitored for merging, with \(|A|\ge2\) for every \(A\in\mathcal{A}\). For each such collection, define its mass and nominal pooled distribution by
\begin{equation}
\label{eq:pooled}
m_A:=\sum_{j\in A}m_j,
\qquad
q_0^A:=\sum_{j\in A}\frac{m_j}{m_A}\,q_0^j.
\end{equation}
At temperature \(T\), we assume that \(q_0^j\) is stable for every
\(j\in\mathcal{S}\), and \(q_0^A\) is unstable for every
\(A\in\mathcal{A}\).
We define the clustering's robustness certificate as its
\(\mathsf{W}_\U\)-distance to the nearest monitored event. This distance
is exactly
\begin{equation}
\label{eq:clustering-certificate}
R_\U(\mathbf{q}_0;T)
=\min\Big\{
\min_{j\in\mathcal{S}}\sqrt{m_j}\,\rho_{\mathrm{s},\U}(q_0^j;T),\;
\min_{A\in\mathcal{A}}\sqrt{m_A}\,\rho_{\mathrm{m},\U}(q_0^A;T)
\Big\},
\end{equation}
since the minimum cost of reaching an event is achieved by moving
only the clusters involved in it. A structured shift that splits cluster \(j\) has \(\mathsf{W}_\U\)-distance from \(\mathbf{q}_0\) of at least
\(\sqrt{m_j}\,\rho_{\mathrm{s},\U}(q_0^j;T)\), with equality for the shift
that moves cluster \(j\) by a plan attaining its split radius (which
exists by Theorem~\ref{thrm:split-sdp}) and leaves the other clusters
unchanged. Likewise, a structured shift that brings the pooled
distribution of \(A\) to or beyond the merge threshold has
\(\mathsf{W}_\U\)-distance of at least
\(\sqrt{m_A}\,\rho_{\mathrm{m},\U}(q_0^A;T)\), with equality, when this
radius is finite, for the shift that moves the clusters in \(A\) by a
plan that attains the merge radius of \(q_0^A\) (which exists by
Proposition~\ref{prop:merge}) and leaves the other clusters unchanged.
Appendix~\ref{app:certificate} gives the formal argument, including the
decomposition of this pooled plan into individual cluster plans.

Consequently, every structured shift with
\(\mathsf{W}_\U(\mathbf{q}_0,\mathbf{q})<R_\U(\mathbf{q}_0;T)\) avoids all
monitored events, and when \(R_\U(\mathbf{q}_0;T)\) is finite, some shift
at this distance reaches at least one of the events. Computing the robustness certificate therefore reduces to solving the split and merge radius problems \eqref{eq:split-radius}--\eqref{eq:merge-radius}, which we
investigate next.


\section{Main Results: Split and Merge Radii and SDP Reformulations}
\label{sec:solution}

The phase-transition analysis of \citet{rose_deterministic_1998} turns
the Hessian conditions of DA stability (Definition~\ref{def:da-stability})
into conditions on the covariance matrix, which we state in our notation as the following lemma. All proofs for this section are given in Appendix~\ref{app:proofs}.

\begin{lemma}[\citet{rose_deterministic_1998}]
\label{lem:da-covariance}
Let \(q\in\mathcal{P}_2(\R^d)\), \(T>0\), and \(k\ge2\). Then
\begin{equation}
\label{eq:covariance-stability}
\nabla_Y^2\mathsf{F}_T(Y_\mu(q);q)\succeq0
\quad\Longleftrightarrow\quad
C(q)\preceq\tfrac{T}{2}I_d,
\end{equation}
with strict positive definiteness of the left-hand side if and only if the inequality on the right-hand side is strict. Hence \(q\) is stable, critical, or unstable depending on whether \(\lambda_{\max}(C(q))\) is less than, equal to, or greater than \(T/2\), independently of \(k\).
\end{lemma}
\begin{proof}
The proof is given in Appendix~\ref{app:da-covariance-stability}.
\end{proof}

We now incorporate the subspace constraint into the displacement
coordinates. Let \(\ell:=\dim\U\), and let
\(B\in\R^{d\times\ell}\) have orthonormal columns spanning \(\U\).
Since columns of \(B\) form a basis for \(\U\), every structured displacement is uniquely represented as \(z-x=B\delta\). We therefore optimize over \emph{displacement plans}, which are the joint distributions of nominal points and their displacement coordinates
\[
\Pi(q_0)
:=
\bigl\{
\pi\in\mathcal{P}_2(\R^d\times\R^\ell):
\pi(\,\cdot\,\times\R^\ell)=q_0
\bigr\}.
\]
For \((X,\Delta)\sim\pi\), let \(q_\pi\) be the distribution of
\(Z=X+B\Delta\). This construction gives a one-to-one correspondence between any \(\U\)-structured transport plan \(\gamma\in\Gamma_\U(q_0,q)\) and its displacement plan \(\pi\in\Pi(q_0)\), with \(q_\pi=q\). The inverse is \(\Delta=B^\top(Z-X)\), and the correspondence preserves cost because \(\|Z-X\|^2=\|\Delta\|^2\).
Using Lemma~\ref{lem:da-covariance}, the split and merge problems
therefore become
\begin{equation}
\label{eq:displacement-problems}
\rho_{e,\U}^2(q_0;T)
=\inf_{\pi\in\Pi(q_0)}
\left\{\int\|\delta\|^2\,\pi(dx,d\delta):\ C(q_\pi)\in\mathcal{E}_e\right\},
\qquad e\in\{\mathrm{s},\mathrm{m}\},
\end{equation}
where \(\mathcal{E}_{\mathrm{s}}:=\{C:\lambda_{\max}(C)\ge T/2\}\) and
\(\mathcal{E}_{\mathrm{m}}:=\{C:C\preceq (T/2)I_d\}\). Without loss of generality, we restrict the problems above to displacement plans with zero displacement mean, since otherwise centering the displacements leaves \(C(q_\pi)\) unchanged and strictly reduces
the cost.

Let \(\mu_0:=\mu(q_0)\) and \(C_0:=C(q_0)\). For
\((X,\Delta)\sim\pi\) with \(\mathbb{E}_\pi[\Delta]=0\), define
\begin{equation}
\label{eq:displacement-moments}
K:=\mathbb{E}_\pi \bigl[(X-\mu_0)\Delta^\top\bigr]
\in\R^{d\times\ell},
\qquad
S:=\mathbb{E}_\pi \bigl[\Delta\Delta^\top\bigr]
\in\R^{\ell\times\ell}
\end{equation}
for the cross-covariance between nominal points and their
displacements and the displacement covariance, respectively. Furthermore, define
\begin{equation}
\label{eq:moment-feasibility}
\begin{aligned}
\mathcal{M}(C_0)
&:=
\left\{
(K,S)\in\R^{d\times\ell}\times\mathbb{S}^{\ell}:
\begin{pmatrix}
C_0 & K\\
K^\top & S
\end{pmatrix}\succeq0
\right\},\\
C_B(K,S)
&:=C_0+KB^\top+BK^\top+BSB^\top .
\end{aligned}
\end{equation}
\begin{proposition}
\label{prop:moment-reduction}
The split and merge radii problems are equivalent to
\begin{align}
\rho_{\mathrm{s},\U}^2(q_0;T)
&=
\inf_{(K,S)\in\mathcal{M}(C_0)}
\bigl\{\operatorname{tr}S:
\lambda_{\max}(C_B(K,S))\ge T/2\bigr\},
\label{eq:split-moment}\\
\rho_{\mathrm{m},\U}^2(q_0;T)
&=
\inf_{(K,S)\in\mathcal{M}(C_0)}
\bigl\{\operatorname{tr}S:
C_B(K,S)\preceq(T/2)I_d\bigr\}.
\label{eq:merge-moment}
\end{align}
Every feasible pair \((K,S)\) is realized by the joint law of
\((X,\Delta)\), where \(X\sim q_0\) and
\(\Delta=K^\top C_0^\dagger(X-\mu_0)+\eta\), where \(\eta\)
is zero mean and independent of \(X\), with covariance
\(S-K^\top C_0^\dagger K\).
If \((K,S)\) is optimal, the joint law of \((X,X+B\Delta)\)
is an optimal \(\U\)-structured transport plan.
\end{proposition}

\begin{proof}
The proof is given in Appendix~\ref{app:moment-reduction}.
\end{proof}

We next provide SDP formulations of both split \eqref{eq:split-moment} and merge \eqref{eq:merge-moment} problems and characterize their feasibility, attainment, and optimal transport plans.

\paragraph{The Merge Problem.}
The merge problem \eqref{eq:merge-moment} is an SDP. The following
proposition characterizes its feasibility and shows that, when feasible,
a deterministic affine map attains the minimum.

\begin{proposition}
\label{prop:merge}
Let \(P_{\U^\perp}:=I_d-BB^\top\) be the orthogonal projection
onto \(\U^\perp\). The merge problem \eqref{eq:merge-moment}
is feasible if and only if
\begin{equation}
\label{eq:merge-feasibility}
P_{\U^\perp}C_0P_{\U^\perp}
\preceq \frac{T}{2}P_{\U^\perp}.
\end{equation}
Whenever feasible, the minimum is attained by a pair
\((K^\star,S^\star)\) satisfying
\(S^\star=(K^\star)^\top C_0^\dagger K^\star\).
The corresponding optimal \(\U\)-structured transport plan
is induced by the deterministic affine map
\(x\mapsto x+B(K^\star)^\top C_0^\dagger(x-\mu_0)\).
\end{proposition}

\begin{proof}
The proof is given in Appendix~\ref{app:merge}.
\end{proof}

\paragraph{The Split Problem.}
The split constraint in \eqref{eq:split-moment} is generally
nonconvex, so the problem is not directly an SDP. It holds exactly
when \(v^\top C_B(K,S)v\ge T/2\) for some unit vector \(v\).
Fixing \(v\) gives an SDP with a closed-form solution for the
smallest shift needed to reach the split threshold along \(v\).
Minimizing this quantity over unit directions yields the
split radius and an optimal split direction.

\begin{lemma}
\label{lem:directional-split}
Let \(q_0\) be stable at temperature \(T>0\), so that
\(C_0\prec(T/2)I_d\). For any unit direction \(v\notin\U^\perp\),
the minimum structured shift needed to reach the split
threshold along \(v\) is
\begin{equation}
\label{eq:directional-split}
\min_{(K,S)\in\mathcal M(C_0)}
\left\{
\operatorname{tr}S:
v^\top C_B(K,S)v\ge\frac{T}{2}
\right\}
=
\frac{\bigl(\sqrt{T/2}-\sqrt{v^\top C_0v}\bigr)^2}{\|B^\top v\|^2}.
\end{equation}
It is feasible exactly when \(v\notin\U^\perp\), and the minimum is attained with \(v^\top C_B(K,S)v=T/2\).
\end{lemma}
\begin{proof}
The proof is given in Appendix~\ref{app:directional-split}.
\end{proof}

Notice that a unit direction \(v\notin\U\) has \(\|B^\top v\|<1\), which increases the directional cost, but a sufficiently small numerator can compensate and make it optimal (Figure~\ref{fig:intro}). In particular, if \(\U=\R^d\), then \(\|B^\top v\|=1\) for every unit \(v\), so the cost is minimized along the top eigenvector of \(C_0\), giving \(\rho_{\mathrm{s},\U}(q_0;T)=\sqrt{T/2}-\sqrt{\lambda_{\max}(C_0)}\). In general, minimizing this cost over unit directions is nonconvex, but the next theorem reformulates it exactly as an SDP.

\begin{theorem}
\label{thrm:split-sdp}
Let \(q_0\) be stable at temperature \(T>0\). Then the squared split radius is given by the following SDP, which is feasible, and its maximum is always attained:
\begin{equation}
\label{eq:split-sdp}
\begin{aligned}
\rho_{\mathrm{s},\U}^{2}(q_0;T)
=\max_{\eta,\tau\ge0}\quad & \eta\\
\textup{subject to}\quad &
\begin{pmatrix}
\tau\bigl(\tfrac{T}{2}I_d-C_0\bigr)-\eta I_d & -\tau C_0^{1/2}B\\
-\tau B^\top C_0^{1/2} & (1-\tau)I_\ell
\end{pmatrix}\succeq0.
\end{aligned}
\end{equation}
\end{theorem}

\begin{proof}
The proof is given in Appendix~\ref{app:split-sdp}, together with the recovery of an attaining split direction and its \(\U\)-structured transport plan.
\end{proof}

\section{Numerical Experiments}
\label{sec:experiments}

\paragraph{S1--S3 benchmarks.}
We use the S1--S3 benchmarks of \citet{franti_iterative_2006}: \(5000\) points in \(\R^2\) from \(15\) Gaussian clusters, with overlap increasing from S1 to S3. Each dataset is centered and normalized to unit root-mean-square norm. We obtain nominal clusterings by running Lloyd's algorithm initialized at the supplied ground-truth centers; we then hold the labels fixed. 

\begin{figure}[!htbp]
\centering
\includegraphics[width=0.9\textwidth]{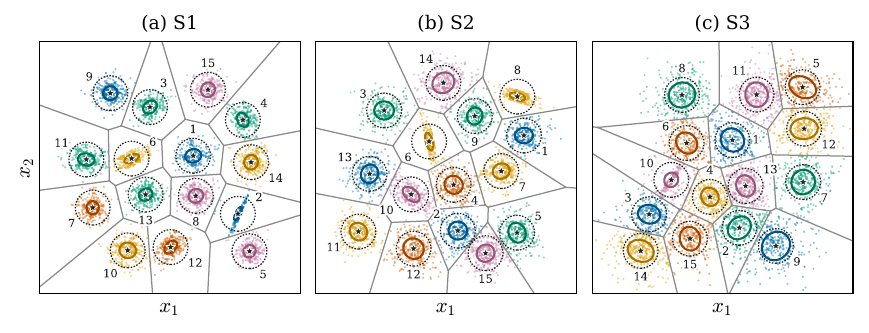}
\caption{Nominal clusterings S1--S3 at \(T=0.1\).}
\label{fig:sipu}
\end{figure}
Figure~\ref{fig:sipu} shows the representatives (stars), their Voronoi cells, the covariance ellipses, with semi-axes \(\sqrt{\lambda_i(C_j)}\), and concentric circles of radius \(\sqrt{T/2}\). By Lemma~\ref{lem:da-covariance}, a cluster is stable when its ellipse lies strictly inside its circle and reaches the
split threshold when the two first touch. For a monitored merge,
the analogous threshold is reached when the pooled covariance
ellipse first fits inside the circle about its pooled mean;
Figure~\ref{fig:binding} of the appendix illustrates attaining merge shifts.

We monitor all \(15\) splits and the merges of neighboring clusters whose Voronoi cells share an edge. We pick \(T=0.1\) from the common range of temperatures (approximately \(0.0759<T<0.1315\)) for which every cluster is stable and the pooled distribution of every monitored pair is unstable in all three datasets. Table~\ref{tab:certificates} reports the certificates and attaining events under unrestricted, horizontal, and vertical shifts (\(\U=\R^2\), \(\operatorname{span}\{e_1\}\), \(\operatorname{span}\{e_2\}\)); all radii in this section are in units of \(10^{-3}\). The radius SDPs are solved with Clarabel through CVXPY. In S1, merge \(\{6,13\}\) is the most vulnerable event under unrestricted shifts, with radius \(12.08\). Its two clusters are aligned nearly vertically, so vertical shifts attain almost the same radius (\(13.12\)), whereas under horizontal shifts the most vulnerable event becomes merge \(\{3,9\}\), at almost twice the radius (\(23.38\)). In S3, split \(9\) is the most vulnerable event under all three choices
of \(\U\) at \(T=0.1\). To examine the effect of resolution, we keep
the S3 clustering and monitored events fixed and consider unrestricted
shifts. As \(T\) increases from \(0.10\) to \(0.12\), the minimum
mass-weighted split radius rises from \(7.23\) to \(12.59\), while the
minimum mass-weighted merge radius falls from \(15.98\) to \(7.90\).
At this coarser resolution, merge \(\{4,13\}\) therefore replaces
split \(9\) as the most vulnerable event. Appendix~\ref{app:separation} gives event rankings, attaining shifts,
and further results for S3 with its clusters moved farther apart.
\begin{table}[!htbp]
\centering
\begin{tabular}{lcrcrcr}
\hline
& \multicolumn{2}{c}{Unrestricted}
& \multicolumn{2}{c}{Horizontal}
& \multicolumn{2}{c}{Vertical} \\
\cline{2-7}
Dataset & Event & \(R_\U\) & Event & \(R_\U\) & Event & \(R_\U\) \\
\hline
S1 & Merge \(\{6,13\}\) & 12.08 & Merge \(\{3,9\}\)  & 23.38 & Merge \(\{6,13\}\) & 13.12 \\
S2 & Merge \(\{2,15\}\) & 12.16 & Merge \(\{2,15\}\) & 16.69 & Merge \(\{1,8\}\)  & 15.49 \\
S3 & Split \(9\)        & 7.23  & Split \(9\)        & 8.96  & Split \(9\)        & 8.54 \\
\hline
\end{tabular}
\caption{Certificates \(R_\U\) (units of \(10^{-3}\)) and attaining events at \(T=0.1\).}
\label{tab:certificates}
\end{table}

\paragraph{Higher-dimensional example: DIM64.}
To demonstrate the SDP formulations in higher dimensions, we consider the DIM64 benchmark from the same repository: \(1024\) points in \(\R^{64}\) from \(16\) clusters. We monitor all \(16\) splits and all \(120\) pairwise merges. Keeping the nominal clustering and \(T=0.1\) fixed, we enlarge the admissible subspace \(\U_\ell=\operatorname{span}\{e_1,\ldots,e_\ell\}\) for \(\ell\in\{10,30,50,55,60,64\}\), with \(\ell=64\) permitting unrestricted shifts. Since the subspaces \(\U_\ell\) are nested, the event radii are nonincreasing in \(\ell\) and the number of feasible merges is nondecreasing. Split \(13\) attains the certificate for every tested \(\ell\), and the
certificate decreases from \(53.6\) to \(50.7\).
No merge is feasible for \(\ell\le 50\), and all \(120\) are feasible
at \(\ell=64\). Table~\ref{tab:dim64} in Appendix~\ref{app:separation} gives the detailed setup and numerical results.

\section{Conclusion and discussion}

We used the phase-transition analysis of deterministic annealing to formalize split and merge thresholds at a prescribed resolution. We quantified the smallest structured distribution shifts reaching these thresholds, which we call the split and merge radii. We reformulated their computation as exact SDPs, and combined them into an exact robustness certificate for a given labeled clustering. This certificate equals the smallest structured population transport distance to any monitored event. Experiments on two-dimensional and higher-dimensional datasets demonstrated the computational tractability of the formulations and the effect of displacement restrictions on robustness.

The framework can be extended to more general distortions, such as Bregman divergences, and more general transport costs. Another direction is to consider a population evolving under underlying dynamics and design controls to steer its covariance toward desired split or merge thresholds or maintain robustness against unwanted ones, connecting the framework to covariance steering.



\bibliographystyle{plainnat}
\bibliography{references_zotero}

@article{franti_iterative_2006,
	title = {Iterative shrinking method for clustering problems},
	volume = {39},
	issn = {0031-3203},
	url = {https://www.sciencedirect.com/science/article/pii/S0031320305003778},
	doi = {10.1016/j.patcog.2005.09.012},
	number = {5},
	urldate = {2026-11-11},
	journal = {Pattern Recognition},
	author = {Fränti, Pasi and Virmajoki, Olli},
	month = may,
	year = {2006},
	pages = {761--775},
}

@article{chen_optimal_2016,
	title = {Optimal {Steering} of a {Linear} {Stochastic} {System} to a {Final} {Probability} {Distribution}, {Part} {I}},
	volume = {61},
	issn = {1558-2523},
	url = {https://ieeexplore.ieee.org/document/7160692},
	doi = {10.1109/TAC.2015.2457784},
	number = {5},
	urldate = {2026-11-09},
	journal = {IEEE Transactions on Automatic Control},
	author = {Chen, Yongxin and Georgiou, Tryphon T. and Pavon, Michele},
	month = may,
	year = {2016},
	pages = {1158--1169},
}

@article{mccann_optimal_2020,
	title = {Optimal {Transportation} {Between} {Unequal} {Dimensions}},
	volume = {238},
	issn = {1432-0673},
	url = {https://doi.org/10.1007/s00205-020-01569-5},
	doi = {10.1007/s00205-020-01569-5},
	language = {en},
	number = {3},
	urldate = {2026-11-03},
	journal = {Archive for Rational Mechanics and Analysis},
	author = {McCann, Robert J. and Pass, Brendan},
	month = dec,
	year = {2020},
	pages = {1475--1520},
}

@inproceedings{guo_low_2020,
	title = {Low {Frequency} {Adversarial} {Perturbation}},
	issn = {2640-3498},
	url = {https://proceedings.mlr.press/v115/guo20a.html},
	language = {en},
	urldate = {2026-11-02},
	booktitle = {Proceedings of {The} 35th {Uncertainty} in {Artificial} {Intelligence} {Conference}},
	publisher = {PMLR},
	author = {Guo, Chuan and Frank, Jared S. and Weinberger, Kilian Q.},
	month = aug,
	year = {2020},
	pages = {1127--1137},
}

@inproceedings{ustun_actionable_2019,
	address = {New York, NY, USA},
	series = {{FAT}* '19},
	title = {Actionable {Recourse} in {Linear} {Classification}},
	isbn = {978-1-4503-6125-5},
	url = {https://dl.acm.org/doi/10.1145/3287560.3287566},
	doi = {10.1145/3287560.3287566},
	urldate = {2026-11-02},
	booktitle = {Proceedings of the {Conference} on {Fairness}, {Accountability}, and {Transparency}},
	publisher = {Association for Computing Machinery},
	author = {Ustun, Berk and Spangher, Alexander and Liu, Yang},
	month = jan,
	year = {2019},
	pages = {10--19},
}

@article{sharma_scalable_2008,
	title = {A {Scalable} {Approach} to {Combinatorial} {Library} {Design} for {Drug} {Discovery}},
	volume = {48},
	issn = {1549-9596},
	url = {https://doi.org/10.1021/ci700023y},
	doi = {10.1021/ci700023y},
	number = {1},
	urldate = {2026-10-30},
	journal = {Journal of Chemical Information and Modeling},
	author = {Sharma, Puneet and Salapaka, Srinivasa and Beck, Carolyn},
	month = jan,
	year = {2008},
	pages = {27--41},
}

@article{gao_distributionally_2023,
	title = {Distributionally {Robust} {Stochastic} {Optimization} with {Wasserstein} {Distance}},
	volume = {48},
	issn = {0364-765X},
	url = {https://pubsonline.informs.org/doi/full/10.1287/moor.2022.1275},
	doi = {10.1287/moor.2022.1275},
	number = {2},
	urldate = {2026-10-15},
	journal = {Mathematics of Operations Research},
	publisher = {INFORMS},
	author = {Gao, Rui and Kleywegt, Anton},
	month = may,
	year = {2023},
	pages = {603--655},
}

@inproceedings{ben-david_sober_2006,
	address = {Berlin, Heidelberg},
	title = {A {Sober} {Look} at {Clustering} {Stability}},
	isbn = {978-3-540-35296-9},
	doi = {10.1007/11776420_4},
	language = {en},
	booktitle = {Learning {Theory}},
	publisher = {Springer},
	author = {Ben-David, Shai and von Luxburg, Ulrike and Pál, Dávid},
	editor = {Lugosi, Gábor and Simon, Hans Ulrich},
	year = {2006},
	pages = {5--19},
}

@article{mavridis_annealing_2023,
	title = {Annealing {Optimization} for {Progressive} {Learning} {With} {Stochastic} {Approximation}},
	volume = {68},
	issn = {1558-2523},
	url = {https://ieeexplore.ieee.org/document/10002308},
	doi = {10.1109/TAC.2022.3232706},
	number = {5},
	urldate = {2026-10-15},
	journal = {IEEE Transactions on Automatic Control},
	author = {Mavridis, Christos N. and Baras, John S.},
	month = may,
	year = {2023},
	pages = {2862--2874},
}

@article{peyre_computational_2019,
	title = {Computational {Optimal} {Transport} with {Applications} to {Data} {Sciences}},
	volume = {11},
	issn = {1935-8237},
	url = {https://doi.org/10.1561/2200000073},
	doi = {10.1561/2200000073},
	number = {5-6},
	urldate = {2026-10-15},
	journal = {Foundations and Trends in Machine Learning},
	author = {Peyré, Gabriel and Cuturi, Marco},
	month = feb,
	year = {2019},
	pages = {355--607},
}

@article{von_luxburg_clustering_2010,
	title = {Clustering {Stability}: {An} {Overview}},
	volume = {2},
	issn = {1935-8237},
	shorttitle = {Clustering {Stability}},
	url = {https://doi.org/10.1561/2200000008},
	doi = {10.1561/2200000008},
	number = {3},
	urldate = {2026-10-15},
	journal = {Foundations and Trends in Machine Learning},
	author = {von Luxburg, Ulrike},
	month = apr,
	year = {2010},
	pages = {235--274},
}

@article{awasthi_center-based_2012,
	title = {Center-based clustering under perturbation stability},
	volume = {112},
	issn = {0020-0190},
	url = {https://www.sciencedirect.com/science/article/pii/S0020019011002778},
	doi = {10.1016/j.ipl.2011.10.006},
	number = {1},
	urldate = {2026-10-15},
	journal = {Information Processing Letters},
	author = {Awasthi, Pranjal and Blum, Avrim and Sheffet, Or},
	month = jan,
	year = {2012},
	pages = {49--54},
}

@article{mohajerin_esfahani_data-driven_2018,
	title = {Data-driven distributionally robust optimization using the {Wasserstein} metric: performance guarantees and tractable reformulations},
	volume = {171},
	issn = {1436-4646},
	shorttitle = {Data-driven distributionally robust optimization using the {Wasserstein} metric},
	url = {https://doi.org/10.1007/s10107-017-1172-1},
	doi = {10.1007/s10107-017-1172-1},
	language = {en},
	number = {1},
	urldate = {2026-10-15},
	journal = {Mathematical Programming},
	author = {Mohajerin Esfahani, Peyman and Kuhn, Daniel},
	month = sep,
	year = {2018},
	pages = {115--166},
}

@misc{balci_exact_2022,
	title = {Exact {SDP} {Formulation} for {Discrete}-{Time} {Covariance} {Steering} with {Wasserstein} {Terminal} {Cost}},
	url = {http://arxiv.org/abs/2205.10740},
	doi = {10.48550/arXiv.2205.10740},
	urldate = {2026-10-15},
	publisher = {arXiv},
	author = {Balci, Isin M. and Bakolas, Efstathios},
	month = may,
	year = {2022},
	note = {arXiv:2205.10740 [math.OC]},
}

@article{fawzi_secure_2014,
	title = {Secure {Estimation} and {Control} for {Cyber}-{Physical} {Systems} {Under} {Adversarial} {Attacks}},
	volume = {59},
	issn = {1558-2523},
	url = {https://ieeexplore.ieee.org/document/6727407},
	doi = {10.1109/TAC.2014.2303233},
	number = {6},
	urldate = {2026-10-15},
	journal = {IEEE Transactions on Automatic Control},
	author = {Fawzi, Hamza and Tabuada, Paulo and Diggavi, Suhas},
	month = jun,
	year = {2014},
	pages = {1454--1467},
}

@inproceedings{bayati_control_2025,
	title = {A {Control} {Barrier} {Function} {Approach} to {Constrained} {Resource} {Allocation} {Problems} in a {Maximum} {Entropy} {Principle} {Framework}},
	issn = {2576-2370},
	url = {https://ieeexplore.ieee.org/abstract/document/11312564},
	doi = {10.1109/CDC57313.2025.11312564},
	urldate = {2026-10-15},
	booktitle = {2025 {IEEE} 64th {Conference} on {Decision} and {Control} ({CDC})},
	author = {Bayati, Alisina and Tiwari, Dhananjay and Salapaka, Srinivasa},
	month = dec,
	year = {2025},
	note = {ISSN: 2576-2370},
	pages = {2481--2486},
}

@article{cortes_coverage_2004,
	title = {Coverage control for mobile sensing networks},
	volume = {20},
	issn = {2374-958X},
	url = {https://ieeexplore.ieee.org/document/1284411},
	doi = {10.1109/TRA.2004.824698},
	number = {2},
	urldate = {2026-10-15},
	journal = {IEEE Transactions on Robotics and Automation},
	author = {Cortes, J. and Martinez, S. and Karatas, T. and Bullo, F.},
	month = apr,
	year = {2004},
	pages = {243--255},
}

@misc{mavridis_multi-resolution_2023,
	title = {Multi-{Resolution} {Online} {Deterministic} {Annealing}: {A} {Hierarchical} and {Progressive} {Learning} {Architecture}},
	shorttitle = {Multi-{Resolution} {Online} {Deterministic} {Annealing}},
	url = {http://arxiv.org/abs/2212.08189},
	doi = {10.48550/arXiv.2212.08189},
	urldate = {2026-09-22},
	publisher = {arXiv},
	author = {Mavridis, Christos and Baras, John},
	month = mar,
	year = {2023},
	note = {arXiv:2212.08189 [cs.LG]},
}

@article{chemseddine_conditional_2025,
	title = {Conditional {Wasserstein} {Distances} with {Applications} in {Bayesian} {OT} {Flow} {Matching}},
	volume = {26},
	issn = {1533-7928},
	url = {http://jmlr.org/papers/v26/24-0586.html},
	number = {141},
	urldate = {2026-09-19},
	journal = {Journal of Machine Learning Research},
	author = {Chemseddine, Jannis and Hagemann, Paul and Steidl, Gabriele and Wald, Christian},
	year = {2025},
	pages = {1--47},
}

@article{srivastava_persistence_2019,
	title = {On the {Persistence} of {Clustering} {Solutions} and {True} {Number} of {Clusters} in a {Dataset}},
	volume = {33},
	copyright = {Copyright (c) 2019 Association for the Advancement of Artificial Intelligence},
	issn = {2374-3468},
	url = {https://ojs.aaai.org/index.php/AAAI/article/view/4431},
	doi = {10.1609/aaai.v33i01.33015000},
	language = {en},
	number = {01},
	urldate = {2026-07-22},
	journal = {Proceedings of the AAAI Conference on Artificial Intelligence},
	author = {Srivastava, Amber and Baranwal, Mayank and Salapaka, Srinivasa},
	month = jul,
	year = {2019},
	pages = {5000--5007},
}

@article{aloise_np-hardness_2009,
	title = {{NP}-hardness of {Euclidean} sum-of-squares clustering},
	volume = {75},
	issn = {1573-0565},
	url = {https://doi.org/10.1007/s10994-009-5103-0},
	doi = {10.1007/s10994-009-5103-0},
	language = {en},
	number = {2},
	urldate = {2026-07-21},
	journal = {Machine Learning},
	author = {Aloise, Daniel and Deshpande, Amit and Hansen, Pierre and Popat, Preyas},
	month = may,
	year = {2009},
	pages = {245--248},
}

@article{rose_deterministic_1998,
	title = {Deterministic annealing for clustering, compression, classification, regression, and related optimization problems},
	volume = {86},
	issn = {1558-2256},
	url = {https://ieeexplore.ieee.org/document/726788/},
	doi = {10.1109/5.726788},
	number = {11},
	urldate = {2026-06-20},
	journal = {Proceedings of the IEEE},
	author = {Rose, K.},
	month = nov,
	year = {1998},
	pages = {2210--2239},
}

@misc{malik_distributionally_2026,
	title = {Distributionally {Robust} {K}-{Means} {Clustering}},
	copyright = {Creative Commons Attribution 4.0 International},
	url = {https://arxiv.org/abs/2604.11118},
	doi = {10.48550/ARXIV.2604.11118},
	language = {en},
	urldate = {2026-06-05},
	publisher = {arXiv},
	author = {Malik, Vikrant and Kargin, Taylan and Hassibi, Babak},
	year = {2026},
	note = {Version Number: 1},
}

\clearpage
\appendix

\section{DA free energy}
\label{app:da-free energy}

Let \(q\in\mathcal{P}_2(\R^d)\),  and let
\(Y=(y_1,\ldots,y_k)\) be a collection of representatives at
fixed \(T>0\).

\paragraph{Entropy formulation.}
Let each point \(x \in \R^d\) be assigned to representative \(y_j\) with
probability \(p_{j|x}\), where
\[
p_{\cdot|x}\in\Delta_k
:=
\left\{p\in[0,1]^k:\sum_{j=1}^k p_j=1\right\}.
\]
For representatives \(Y=(y_1,\ldots,y_k)\), define the expected
distortion and conditional assignment entropy by
\[
\mathcal{D}(Y,p;q)
:=
\int\sum_{j=1}^k p_{j|x}\|x-y_j\|^2\,q(dx),
\qquad
\mathcal{H}(p;q)
:=
-\int\sum_{j=1}^k p_{j|x}\log p_{j|x}\,q(dx),
\]
with \(0\log0=0\). At temperature \(T>0\), the free energy is
\begin{equation}
\label{eq:da-objective}
\begin{aligned}
\mathcal{F}_T(Y,p;q)
&:=\mathcal{D}(Y,p;q)-T\mathcal{H}(p;q)\\
&=\int\sum_{j=1}^k p_{j|x}
\bigl(\|x-y_j\|^2+T\log p_{j|x}\bigr)\,q(dx).
\end{aligned}
\end{equation}
For fixed \(Y\), minimizing over assignments gives the Gibbs probabilities
\begin{equation}
\label{eq:gibbs-assignments}
p^\star_{j|x}(Y)
=
\frac{e^{-\|x-y_j\|^2/T}}
{\sum_{\ell=1}^k e^{-\|x-y_\ell\|^2/T}}.
\end{equation}
Substituting these assignments into \eqref{eq:da-objective} gives
the reduced free energy
\[
\mathsf{F}_T(Y;q)
:=
\inf_p \mathcal{F}_T(Y,p;q)
=
-T\int\log\!\left(
\sum_{j=1}^k e^{-\|x-y_j\|^2/T}
\right)q(dx),
\]
as stated in \eqref{eq:free-energy}.

\paragraph{Zero-temperature limit.}
Let \(m(x):=\min_j\|x-y_j\|^2\). Since
\[
e^{-m(x)/T}
\le
\sum_{j=1}^k e^{-\|x-y_j\|^2/T}
\le
k e^{-m(x)/T},
\]
taking \(-T\log\) and integrating gives
\begin{equation}
\label{eq:free-energy-limit}
\mathcal{D}(Y;q)-T\log k
\le
\mathsf{F}_T(Y;q)
\le
\mathcal{D}(Y;q).
\end{equation}
Thus, since \(k\) is fixed, the reduced free energy converges to the distortion uniformly in \(Y\), as \(T\downarrow0\).

\section{Finiteness and attainment of \(W_{2,\U}\)}
\label{app:conditional-wasserstein}

\begin{lemma}
\label{lem:finiteness}
For \(q_0,q\in\mathcal{P}_2(\R^d)\), \(W_{2,\U}(q_0,q)<\infty\) if
and only if \(q_0\) and \(q\) have the same marginal on \(\U^\perp\),
in which case the infimum in \eqref{eq:restricted-wasserstein} is
attained.
\end{lemma}

\begin{proof}
Let \(P_{\U}\) and \(P_{\U^\perp}\) be the orthogonal projections
onto \(\U\) and \(\U^\perp\). For \(X\sim q_0\), write
\(X_{\U}:=P_{\U}X\) and \(X_{\U^\perp}:=P_{\U^\perp}X\), and
define \(Z_{\U},Z_{\U^\perp}\) similarly for \(Z\sim q\).
Any \(\U\)-structured plan satisfies \(Z-X\in\U\), equivalently 
\(Z_{\U^\perp}=X_{\U^\perp}\), almost surely. Thus, the
\(\U^\perp\)-marginals of \(q\) and \(q_0\) must be equal.

Conversely, suppose that \(X_{\U^\perp}\) and \(Z_{\U^\perp}\)
have the same distribution \(\nu\), and let \(\Xi\sim\nu\)
denote the unchanged \(\U^\perp\) component. Define
\[
\begin{aligned}
P_{\Xi,X_{\U}}:=P_{X_{\U^\perp},X_{\U}},
\qquad
P_{\Xi,Z_{\U}}:=P_{Z_{\U^\perp},Z_{\U}}.
\end{aligned}
\]
These joint distributions share the marginal \(\nu\) on \(\Xi\).

This matches the conditional Wasserstein distance formulation of
\citet{chemseddine_conditional_2025}, with \(\Xi\) as the
conditioning variable. The orthogonal decomposition preserves
cost, since
\[
\|Z-X\|^2
=
\|(Z_{\U^\perp},Z_{\U})-(X_{\U^\perp},X_{\U})\|^2.
\]
Thus, \(W_{2,\U}(q_0,q)\) is equivalent to their conditional
\(2\)-Wasserstein distance between \(P_{\Xi,X_{\U}}\) and
\(P_{\Xi,Z_{\U}}\). Their Proposition~1(iii) constructs an optimal conditional coupling \(\alpha\) between \(P_{\Xi,X_{\U}}\) and
\(P_{\Xi,Z_{\U}}\), preserving the \(\Xi\)-component.
Let \((\Xi_1,V_1,\Xi_2,V_2)\sim\alpha\).
Then the law of \((\Xi_1+V_1,\Xi_2+V_2)\) is an optimal
\(\U\)-structured plan between \(q_0\) and \(q\), since
\(\Xi_1=\Xi_2\) almost surely and this correspondence preserves
transport costs. Its cost is finite because
\(\|z-x\|^2\le 2\|x\|^2+2\|z\|^2\) and
\(q_0,q\in\mathcal{P}_2(\R^d)\).
\end{proof}

\section{Proof of the clustering certificate}
\label{app:certificate}

For a split of cluster \(j\), any shift that brings \(q_0^j\) to or beyond
the split threshold satisfies
\(\mathsf{W}_\U^2(\mathbf q_0,\mathbf q)\ge m_jW_{2,\U}^2(q_0^j,q^j)\ge
m_j\rho_{\mathrm{s},\U}^2(q_0^j;T)\), with equality when \(q_0^j\) is
transported by a plan attaining its split radius and every other cluster
is unchanged.

For a merge, however, it is less obvious that transporting a collection
\(A\) by a pooled structured plan attaining its merge radius
\(\rho_{\mathrm{m},\U}(q_0^A;T)\) incurs \(\mathsf{W}_\U\)-distance
\(\sqrt{m_A}\,\rho_{\mathrm{m},\U}(q_0^A;T)\), since \(\mathsf{W}_\U\)
measures the transport of each cluster separately rather than of the
pooled distribution. To show this, suppose the merge radius of \(q_0^A\)
is finite, and let \(\gamma^{A,\star}\in\Gamma_\U(q_0^A,q^A)\) be such a
plan, i.e., \(q^A\) is at or beyond the merge threshold and
\begin{equation}
\label{eq:pooled-optimal}
\int\|z-x\|^2\,\gamma^{A,\star}(dx,dz)
=W_{2,\U}^2(q_0^A,q^A)
=\rho_{\mathrm{m},\U}^2(q_0^A;T).
\end{equation}
We can condition this pooled plan on the cluster label to obtain a
\(\U\)-structured transport plan \(\gamma^j\) for each cluster \(j\in A\),
whose first marginal is \(q_0^j\) by construction. Denote the second marginal by \(q^j\), such that
\begin{equation}
\label{eq:pooled-decomposition}
\gamma^{A,\star}=\sum_{j\in A}\frac{m_j}{m_A}\,\gamma^j,
\qquad
q^A=\sum_{j\in A}\frac{m_j}{m_A}\,q^j.
\end{equation}
Each \(\gamma^j\) is an optimal structured plan between \(q_0^j\) and
\(q^j\). Otherwise, some \(\widetilde\gamma^j\in\Gamma_\U(q_0^j,q^j)\) has
a lower cost, and replacing \(\gamma^j\) by \(\widetilde\gamma^j\) in
\eqref{eq:pooled-decomposition} gives a plan in \(\Gamma_\U(q_0^A,q^A)\)
with a lower cost than \(\gamma^{A,\star}\), contradicting its
optimality. Plugging \eqref{eq:pooled-decomposition} into
\eqref{eq:pooled-optimal} therefore gives
\[
m_A\rho_{\mathrm{m},\U}^2(q_0^A;T)
=\sum_{j\in A}m_j\int\|z-x\|^2\,\gamma^j(dx,dz)
=\sum_{j\in A}m_jW_{2,\U}^2(q_0^j,q^j).
\]
Transporting each cluster \(j\in A\) by \(\gamma^j\) and leaving every
other cluster unchanged thus reaches the merge event at
\(\mathsf{W}_\U\)-distance \(\sqrt{m_A}\,\rho_{\mathrm{m},\U}(q_0^A;T)\).

\section{Proofs}
\label{app:proofs}

\subsection{Second variation and DA stability}
\label{app:da-covariance-stability}

Let \(q\in\mathcal{P}_2(\R^d)\), \(T>0\), and \(k\ge2\).
Write \(\mu:=\mu(q)\), \(C:=C(q)\), and
\(Y_\mu:=Y_\mu(q)=(\mu,\ldots,\mu)\).
We first derive the second variation at \(Y_\mu\), then use it
to prove Lemma~\ref{lem:da-covariance}.

\paragraph{Second variation at coincidence.}
For a perturbation \(\phi=(\phi_1,\ldots,\phi_k)\), with
\(\phi_j\in\R^d\), define
\[
y_j^\varepsilon:=\mu+\varepsilon\phi_j,
\qquad
Y^\varepsilon:=Y_\mu+\varepsilon\phi.
\]

\begin{lemma}[Second variation]
\label{lem:da-second-variation}
The configuration \(Y_\mu\) is stationary, and the second variation
of \(\mathsf{F}_T\) at \(Y_\mu\) is
\begin{align}
\label{eq:da-second-variation}
Q(\phi)
&:=
\left.
\frac{d^2}{d\varepsilon^2}
\mathsf{F}_T(Y^\varepsilon;q)
\right|_{\varepsilon=0}
\nonumber\\
&=
\frac{4}{T}
\int
\left(
\frac{1}{k}\sum_{j=1}^k
\phi_j^\top(x-\mu)
\right)^2q(dx)
+
\frac{2}{k}\sum_{j=1}^k
\phi_j^\top
\left(I_d-\frac{2}{T}C\right)\phi_j.
\end{align}
\end{lemma}

\begin{proof}
Write \(p^\varepsilon_{j|x}:=p^\star_{j|x}(Y^\varepsilon)\)
for the Gibbs assignments at the perturbed configuration.
Differentiating \eqref{eq:free-energy} with respect to
\(\varepsilon\) gives
\begin{equation}
\label{eq:da-first-variation}
\frac{d}{d\varepsilon}\mathsf{F}_T(Y^\varepsilon;q)
=
-2\int
\sum_{j=1}^k
p^\varepsilon_{j|x}\,
\phi_j^\top(x-\mu-\varepsilon\phi_j)\,q(dx),
\end{equation}
where differentiation under the integral, for both first and second variations, is justified by the finite second moment of \(q\).

At \(\varepsilon=0\), the assignments are uniform,
\(p^0_{j|x}=1/k\), so
\[
\left.
\frac{d}{d\varepsilon}\mathsf{F}_T(Y^\varepsilon;q)
\right|_{\varepsilon=0}
=
-\frac{2}{k}\sum_{j=1}^k
\phi_j^\top\int(x-\mu)\,q(dx)
=
0.
\]
Since this holds for every \(\phi\),
\(\nabla_Y\mathsf{F}_T(Y_\mu;q)=0\).

Differentiating the Gibbs assignments gives
\begin{equation}
\label{eq:da-assignment-variation}
\frac{d}{d\varepsilon}p^\varepsilon_{j|x}
=
\frac{2}{T}p^\varepsilon_{j|x}
\left[
\phi_j^\top(x-\mu-\varepsilon\phi_j)
-
\sum_{\ell=1}^k
p^\varepsilon_{\ell|x}
\phi_\ell^\top(x-\mu-\varepsilon\phi_\ell)
\right].
\end{equation}
Differentiating \eqref{eq:da-first-variation} once more, using
\eqref{eq:da-assignment-variation}, and evaluating at
\(\varepsilon=0\) yields
\begin{align*}
Q(\phi)
&=
\frac{2}{k}
\int\sum_{j=1}^k\|\phi_j\|^2\,q(dx)
\\
&\quad-
\frac{4}{kT}
\int\sum_{j=1}^k
\bigl(\phi_j^\top(x-\mu)\bigr)^2\,q(dx)
\\
&\quad+
\frac{4}{T}
\int
\left(
\frac{1}{k}\sum_{j=1}^k
\phi_j^\top(x-\mu)
\right)^2q(dx).
\end{align*}
By the definition of \(C\), the first two terms equal
\[
\frac{2}{k}\sum_{j=1}^k
\phi_j^\top
\left(I_d-\frac{2}{T}C\right)\phi_j,
\]
which gives \eqref{eq:da-second-variation}.
\end{proof}

\paragraph{Covariance criterion.}
We now establish the stability condition from the second
variation.

\begin{proof}[Proof of Lemma~\ref{lem:da-covariance}]
The first term of \eqref{eq:da-second-variation} is nonnegative.
If \(\lambda_{\max}(C)<T/2\), then
\[
I_d-\frac{2}{T}C\succ0.
\]
The second term is therefore strictly positive for every nonzero
\(\phi\). Thus, the Hessian is positive definite, and stationarity
implies that \(Y_\mu\) is a strict local minimum.

Let \(v\) be a unit eigenvector corresponding to
\(\lambda_{\max}(C)\), and define
\[
\varphi_j
=
\begin{cases}
v, & j=1,\\
-v, & j=2,\\
0, & j\ge3.
\end{cases}
\]
Because \(\sum_j\varphi_j=0\), the first term of
\eqref{eq:da-second-variation} vanishes, giving
\[
Q(\varphi)
=
\frac{4}{k}
v^\top\left(I_d-\frac{2}{T}C\right)v
=
\frac{4}{k}
\left(1-\frac{2\lambda_{\max}(C)}{T}\right).
\]
If \(\lambda_{\max}(C)>T/2\), then \(Q(\varphi)<0\), so the
Hessian has a negative eigenvalue and \(Y_\mu\) is a saddle point.

Finally, if \(\lambda_{\max}(C)=T/2\), then
\(I_d-2C/T\succeq0\), so \(Q(\phi)\ge0\) for every \(\phi\).
The nonzero perturbation \(\varphi\) satisfies \(Q(\varphi)=0\).
Thus, the Hessian is positive semidefinite but singular, and
\(q\) is critical.  These three cases establish the claimed equivalences for every \(k\ge2\).
\end{proof}

\subsection{Proof of Proposition~\ref{prop:moment-reduction}}
\label{app:moment-reduction}

Let \(\pi\in\Pi(q_0)\) and \((X,\Delta)\sim\pi\). As discussed in the main text, we assume without loss of generality that \(\mathbb{E}_\pi[\Delta]=0\). Let \(K\) and \(S\) be the associated
moment matrices defined in \eqref{eq:displacement-moments}.
The covariance of the stacked random vector
\((X^\top,\Delta^\top)^\top\in\R^{d+\ell}\) is given by
\[
\begin{pmatrix}
C_0 & K\\
K^\top & S
\end{pmatrix}
=
\mathbb{E}_\pi \left[
\begin{pmatrix}
X-\mu_0\\
\Delta
\end{pmatrix}
\begin{pmatrix}
X-\mu_0\\
\Delta
\end{pmatrix}^{\!\top}
\right]
\succeq0,
\]
so \((K,S)\in\mathcal{M}(C_0)\). Moreover,
\[
C(q_\pi)
=
C_0+KB^\top+BK^\top+BSB^\top
=
C_B(K,S),
\qquad
\mathbb{E}_\pi\left[\|\Delta\|^2\right]=\operatorname{tr}S.
\]
Hence, every centered displacement plan yields a feasible moment
pair with the same cost and target covariance.

Conversely, let \((K,S)\in\mathcal{M}(C_0)\). The generalized
Schur complement gives
\[
\qquad
V:=S-K^\top C_0^\dagger K\succeq0,
\qquad
C_0C_0^\dagger K=K.
\]
Choose \(X\sim q_0\) and a zero-mean random vector \(\eta\in\R^\ell\),
independent of \(X\), with covariance \(V\) (for example
\(\eta\sim\mathcal{N}(0,V)\)), and set
\[
\Delta:=K^\top C_0^\dagger(X-\mu_0)+\eta.
\]
The joint law \(\pi\) of \((X,\Delta)\) belongs to \(\Pi(q_0)\),
and \(\mathbb{E}_\pi[\Delta]=0\). Since \(X-\mu_0\) and \(\eta\) are independent and have zero mean,
\(\mathbb{E}_\pi[(X-\mu_0)\eta^\top]=0\), so
\[
\mathbb{E}_\pi\bigl[(X-\mu_0)\Delta^\top\bigr]
=
C_0C_0^\dagger K
=
K.
\]
Moreover,
\[
\mathbb{E}_\pi\bigl[\Delta\Delta^\top\bigr]
=
K^\top C_0^\dagger C_0C_0^\dagger K+V
=
S.
\]
Thus, every feasible moment pair is realized by a centered
displacement plan.

Finally, if the moment infimum is attained at
\((K^\star,S^\star)\), the construction above yields an optimal
displacement plan \(\pi^\star\). The corresponding optimal
\(\U\)-structured transport plan \(\gamma^\star\) is then given by the joint law of \((X,X+B\Delta)\), where \((X,\Delta)\sim\pi^\star\).

\subsection{Proof of Proposition~\ref{prop:merge}}
\label{app:merge}

\begin{proof}
Since \(P_{\U^\perp} B=0\), multiplying the merge constraint \(C_B(K,S)\preceq \frac{T}{2}I_d\) by \(P_{\U^\perp}\) on both sides gives
\[
P_{\U^\perp} C_0P_{\U^\perp}
=
P_{\U^\perp} C_B(K,S)P_{\U^\perp}
\preceq \frac{T}{2}P_{\U^\perp},
\]
which proves the necessity of \eqref{eq:merge-feasibility}. Conversely, suppose this condition holds and choose
\(K=-C_0B\) and \(S=B^\top C_0B\). Then
\[
\begin{pmatrix}
C_0&K\\ K^\top&S
\end{pmatrix}
=
\begin{pmatrix}
I_d\\ -B^\top
\end{pmatrix}
C_0
\begin{pmatrix}
I_d&-B
\end{pmatrix}
\succeq0,
\]
and
\[
C_B(K,S)
=P_{\U^\perp} C_0P_{\U^\perp}
\preceq \frac{T}{2}P_{\U^\perp}
\preceq \frac{T}{2}I_d.
\]
Thus, the merge problem is feasible.

We next show that the feasible set is compact. For any feasible pair \((K,S)\), multiplying the merge constraint
on the left by \(B^\top\) and on the right by \(B\), and using
\(B^\top B=I_\ell\), gives
\[
S\preceq \frac{T}{2}I_\ell-B^\top C_0B
-\bigl(B^\top K+K^\top B\bigr).
\]
The block PSD constraint, evaluated on \((2Bv,v)\) for every
\(v\in\R^\ell\), yields
\[
B^\top K+K^\top B
\succeq -2B^\top C_0B-\frac12S.
\]
Substituting this bound gives
\(S\preceq \frac{T}{2}I_\ell+B^\top C_0B+\frac12S\).
Rearranging, we obtain
\[
0\preceq S\preceq 2B^\top C_0B+T I_\ell.
\]
Thus, \(S\) is bounded.

The block PSD constraint also implies that its \(2\times2\) principal minors are nonnegative, so
\[
(C_0)_{ii}S_{jj}-K_{ij}^2\ge0,
\qquad i=1,\ldots,d,\quad j=1,\ldots,\ell.
\]
Since \(C_0\) is fixed and \(S\) is bounded, every entry of \(K\)
is bounded, and hence \(K\) is also bounded. The feasible set is also closed and therefore compact, whenever nonempty. Hence, the continuous objective \(\operatorname{tr}S\) attains its minimum.

Finally, let \((K^\star,S^\star)\) be an optimal pair. The
generalized Schur complement gives
\[
C_0C_0^\dagger K^\star=K^\star,
\qquad
V:=S^\star-(K^\star)^\top C_0^\dagger K^\star\succeq0.
\]
Replacing \(S^\star\) by
\(\widetilde S:=(K^\star)^\top C_0^\dagger K^\star\)
still satisfies the block PSD constraint and gives
\[
C_B(K^\star,\widetilde S)
=C_B(K^\star,S^\star)-BVB^\top
\preceq \frac{T}{2}I_d.
\]
Thus, the pair \((K^\star,\widetilde S)\) is feasible and reduces the objective by \(\operatorname{tr}V\). Optimality of \(S^\star\) implies
\(\operatorname{tr}V=0\), and hence \(V=0\).

Consequently, the construction in
Proposition~\ref{prop:moment-reduction} gives an optimal displacement
\[
\Delta=(K^\star)^\top C_0^\dagger(X-\mu_0).
\]
The corresponding optimal \(\U\)-structured transport plan is then given by
\[
x \mapsto x + B (K^\star)^\top C_0^\dagger(x-\mu_0).
\]
\end{proof}

\subsection{Proof of Lemma~\ref{lem:directional-split}}
\label{app:directional-split}

\begin{proof}
Since \(q_0\) is stable, \(v^\top C_0v<T/2\) for every unit
direction \(v\). If \(v\in\U^\perp\), then \(B^\top v=0\), so
\(v^\top C_B(K,S)v=v^\top C_0v<T/2\) for every feasible
\((K,S)\). Therefore, the directional constraint is infeasible.

Now suppose \(B^\top v\ne0\). By
Proposition~\ref{prop:moment-reduction}, every
\((K,S)\in\mathcal M(C_0)\) is realized by a displacement plan
\(\pi\), with \(X\sim q_0\) and \(\mathbb E_\pi[\Delta]=0\).
Expanding the directional constraint gives
\[
v^\top C_0v
+2v^\top KB^\top v
+v^\top BSB^\top v
\ge\frac{T}{2}.
\]
Using the definition of \(K\) and Cauchy--Schwarz, we obtain
\begin{equation}
\label{eq:directional-cross-bound}
\begin{aligned}
v^\top KB^\top v
&=
v^\top\mathbb E_\pi\!\left[
(X-\mu_0)\Delta^\top
\right]B^\top v\\
&=
\mathbb E_\pi\!\left[
\bigl(v^\top(X-\mu_0)\bigr)
\bigl(\Delta^\top B^\top v\bigr)
\right]\\
&\le
\sqrt{\mathbb E_\pi\!\left[
\bigl(v^\top(X-\mu_0)\bigr)^2
\right]}\,
\sqrt{\mathbb E_\pi\!\left[
\bigl(\Delta^\top B^\top v\bigr)^2
\right]}\\
&=
\sqrt{v^\top C_0v}\,
\sqrt{v^\top BSB^\top v}.
\end{aligned}
\end{equation}
It follows that
\[
\begin{aligned}
v^\top C_B(K,S)v
&\le
v^\top C_0v
+2\sqrt{v^\top C_0v}\sqrt{v^\top BSB^\top v}
+v^\top BSB^\top v\\
&=
\left(
\sqrt{v^\top C_0v}
+\sqrt{v^\top BSB^\top v}
\right)^2.
\end{aligned}
\]
Moreover, since \(S=\mathbb E_\pi[\Delta\Delta^\top]\), one more Cauchy--Schwarz gives
\begin{equation}
\label{eq:directional-displacement-bound}
\begin{aligned}
v^\top BSB^\top v
&=v^\top B\,\mathbb E_\pi[\Delta\Delta^\top]B^\top v\\
&=\mathbb E_\pi\!\left[\bigl(\Delta^\top B^\top v\bigr)^2\right]
\le \|B^\top v\|^2\,\mathbb E_\pi\|\Delta\|^2 =\operatorname{tr}S\,\|B^\top v\|^2.
\end{aligned}
\end{equation}
Combining these bounds gives
\begin{equation}
\label{eq:directional-variance-bound}
v^\top C_B(K,S)v
\le
\left(
\sqrt{v^\top C_0v}
+\|B^\top v\|\sqrt{\operatorname{tr}S}
\right)^2.
\end{equation}
The bound \eqref{eq:directional-variance-bound} holds with equality
when both \eqref{eq:directional-cross-bound} and
\eqref{eq:directional-displacement-bound} hold with equality.
For \(\alpha\ge0\), this occurs when
\(\Delta=\alpha B^\top v\,v^\top(X-\mu_0)\) if \(v^\top C_0v>0\),
which gives
\[
K=\alpha C_0v\,v^\top B,
\qquad
S=\alpha^2(v^\top C_0v)B^\top v\,v^\top B.
\]
If \(v^\top C_0v=0\), choose
\(\Delta=\alpha  \kappa B^\top v\), where \(\kappa\) is a zero mean and unit variance random variable independent of \(X\). This gives
\[
K=0,
\qquad
S=\alpha^2B^\top v\,v^\top B.
\]
Rearranging \eqref{eq:directional-variance-bound} under the
constraint \(v^\top C_B(K,S)v\ge T/2\) gives
\begin{equation}
\label{eq:directional-cost-bound}
\operatorname{tr}S
\ge
\frac{\left(\sqrt{T/2}-\sqrt{v^\top C_0v}\right)^2}
{\|B^\top v\|^2}.
\end{equation}
Choosing \(\alpha\) in the above feasible pairs so that
\(v^\top C_B(K,S)v=T/2\) attains equality in
\eqref{eq:directional-cost-bound}. Thus, this lower bound is
attained and equals the minimum structured transport cost needed to reach the split threshold along \(v\).
\end{proof}

\subsection{Proof of Theorem~\ref{thrm:split-sdp} and recovery of a split direction}
\label{app:split-sdp}

\begin{proof}
The problem we want to solve is
\[
\rho_{\mathrm{s},\U}^{\,2}(q_0;T)
=
\min_{\substack{\|v\|=1\\B^\top v\ne0}}
\frac{\left(\sqrt{T/2}-\sqrt{v^\top C_0v}\right)^2}
{\|B^\top v\|^2}.
\]
To put the problem in a form suitable for applying the S-lemma,
we reformulate it in the following way
\[
\begin{aligned}
\rho_{\mathrm{s},\U}^{\,2}(q_0;T)
={}&\max_{\eta\ge0}\ \eta\\
&\text{subject to}\quad
\eta\le
\frac{\left(\sqrt{T/2}-\sqrt{v^\top C_0v}\right)^2}
{\|B^\top v\|^2}
\quad
\text{for all }\|v\|=1,\ B^\top v\ne0.
\end{aligned}
\]
Since \(q_0\) is stable,
\(\sqrt{T/2}-\sqrt{v^\top C_0v}>0\). Taking the square root and rearranging gives
\[
\sqrt{v^\top C_0v}
+\sqrt{\eta}\,\|B^\top v\|
\le\sqrt{T/2}.
\]
This inequality is automatically satisfied when \(B^\top v=0\),
so we can include all unit directions. Hence,
\begin{equation}
\label{eq:split-common-bound}
\rho_{\mathrm{s},\U}^{\,2}(q_0;T)
=
\max_{\eta\ge0}
\left\{
\eta:
\max_{\|v\|=1}
\left(
\sqrt{v^\top C_0v}
+\sqrt{\eta}\,\|B^\top v\|
\right)
\le\sqrt{T/2}
\right\}.
\end{equation}
We next rewrite the inner maximum using the identity
\(\|a\|=\max_{\|u\|=1}u^\top a\). In particular,
\[
\sqrt{v^\top C_0v}
=
\|C_0^{1/2}v\|
=
\max_{\|u\|=1}v^\top C_0^{1/2}u,
\qquad
\sqrt{\eta}\,\|B^\top v\|
=
\max_{\|w\|^2\le\eta}v^\top Bw,
\]
where \(u\in\R^d\) and \(w\in\R^\ell\) are auxiliary vectors.
Their choices are independent, so
\begin{equation}
\label{eq:split-norm-reformulation}
\begin{aligned}
\max_{\|v\|=1}
\left(
\sqrt{v^\top C_0v}
+\sqrt{\eta}\,\|B^\top v\|
\right)
&=
\max_{\|v\|=1}
\max_{\substack{\|u\|=1\\\|w\|^2\le\eta}}
v^\top(C_0^{1/2}u+Bw)\\
&\quad=
\max_{\substack{\|u\|=1\\\|w\|^2\le\eta}}
\max_{\|v\|=1}
v^\top(C_0^{1/2}u+Bw)\\
&\quad=
\max_{\substack{\|u\|=1\\\|w\|^2\le\eta}}
\|C_0^{1/2}u+Bw\|.
\end{aligned}
\end{equation}
Consequently, the constraint in \eqref{eq:split-common-bound}
is equivalent to
\begin{equation}
\label{eq:split-unit-condition}
\|C_0^{1/2}u+Bw\|^2\le\frac{T}{2}
\qquad
\text{whenever }\|u\|=1,\ \|w\|^2\le\eta.
\end{equation}
This is equivalent to the quadratic implication
\begin{equation}
\label{eq:split-quadratic-implication}
\|w\|^2\le\eta\|u\|^2
\quad\Longrightarrow\quad
\|C_0^{1/2}u+Bw\|^2
\le\frac{T}{2}\|u\|^2
\end{equation}
for all \(u,w\). To see this, assume
\eqref{eq:split-unit-condition} holds. For any \(u\ne0\) and
\(w\) satisfying \(\|w\|^2\le\eta\|u\|^2\), apply it to
\(u/\|u\|\) and \(w/\|u\|\), then multiply the resulting
inequality by \(\|u\|^2\). If \(u=0\), then \(w=0\), so the conclusion holds automatically. Conversely,
restricting \eqref{eq:split-quadratic-implication} to
\(\|u\|=1\) recovers \eqref{eq:split-unit-condition}.

For \(\eta>0\),
any \(u\ne0\) and \(w=0\) satisfy
\(\eta\|u\|^2-\|w\|^2>0\).
Therefore, by the S-lemma,
\eqref{eq:split-quadratic-implication} is equivalent to the existence of \(\lambda\ge0\) such that
\begin{equation}
\label{eq:split-slemma-certificate}
\frac{T}{2}\|u\|^2
-\|C_0^{1/2}u+Bw\|^2
-\lambda\bigl(\eta\|u\|^2-\|w\|^2\bigr)
\ge0
\qquad\text{for all }u,w.
\end{equation}
Expanding \(\|C_0^{1/2}u+Bw\|^2 =
u^\top C_0u +2u^\top C_0^{1/2}Bw +w^\top w\), the left-hand side of
\eqref{eq:split-slemma-certificate} equals
\[
\begin{pmatrix}u\\w\end{pmatrix}^{\!\top}
\begin{pmatrix}
\left(\frac{T}{2}-\lambda\eta\right)I_d-C_0
&-C_0^{1/2}B\\
-B^\top C_0^{1/2}
&(\lambda-1)I_\ell
\end{pmatrix}
\begin{pmatrix}u\\w\end{pmatrix}.
\]
Its nonnegativity for every \(u,w\) is equivalent to
\begin{equation}
\label{eq:split-multiplier-lmi}
M(\eta, \lambda)
=
\begin{pmatrix}
\left(\frac{T}{2}-\lambda\eta\right)I_d-C_0
&-C_0^{1/2}B\\
-B^\top C_0^{1/2}
&(\lambda-1)I_\ell
\end{pmatrix}\succeq0.
\end{equation}

Since \(\lambda\ge1\) (by lower-right block), setting \(\tau:=1/\lambda\in(0,1]\) and multiplying
\eqref{eq:split-multiplier-lmi} by \(\tau>0\) gives
\begin{equation}
\label{eq:split-scaled-lmi}
\begin{pmatrix}
\tau\left(\frac{T}{2}I_d-C_0\right)-\eta I_d
&-\tau C_0^{1/2}B\\
-\tau B^\top C_0^{1/2}
&(1-\tau)I_\ell
\end{pmatrix}\succeq0.
\end{equation}
This is now affine in \(\tau, \eta\). For positive \(\eta\), every step above is an equivalence. 

Note that allowing \(\tau=0\) forces \(\eta=0\), which already satisfies
\eqref{eq:split-common-bound} and therefore does not change the optimal value. Hence, maximizing \(\eta\) over
\(\eta,\tau\ge0\) subject to \eqref{eq:split-scaled-lmi}
gives the SDP in \eqref{eq:split-sdp}.

It is straightforward to check that \(0\le\tau\le1\) and \(0\le\eta\le T/2\), and the feasible set is nonempty, so the maximum is attained.
\end{proof}

\paragraph{Recovering an optimal split direction.}
One way to recover an optimal split direction from
\((\eta^\star,\tau^\star)\) is as follows. The constraint
\eqref{eq:split-unit-condition} can be written as
\[
\max_{\substack{\|u\|=1\\\|w\|^2\le\eta^\star}}
\|C_0^{1/2}u+Bw\|^2
\le\frac{T}{2}.
\]
The maximum is attained by compactness. Since the objective is
strictly convex in \(w\), an interior point cannot maximize it,
so any maximizing pair satisfies \(\|w^\star\|^2=\eta^\star\).
Moreover, the inequality must be tight at optimality. Otherwise,
a positive gap would allow every feasible \(w\) to be
scaled by a factor \(\alpha>1\) sufficiently close to one while still satisfying the inequality. The larger budget \(\alpha^2\eta^\star\)
would then be feasible, contradicting optimality.
Thus, an attaining pair satisfies
\[
\|u^\star\|=1,\qquad
\|w^\star\|^2=\eta^\star,\qquad
\|C_0^{1/2}u^\star+Bw^\star\|^2=\frac{T}{2}.
\]

Since \(\eta^\star>0\), we have \(\tau^\star>0\).
Set \(\lambda^\star:=1/\tau^\star\), and let \(M^\star\)
denote the matrix in \eqref{eq:split-multiplier-lmi} evaluated
at \((\eta^\star,\lambda^\star)\).
The equalities above make the S-lemma certificate
\eqref{eq:split-slemma-certificate} zero, giving
\[
\begin{pmatrix}u^\star\\w^\star\end{pmatrix}^{\!\top}
M^\star
\begin{pmatrix}u^\star\\w^\star\end{pmatrix}=0.
\]
Since \(M^\star\succeq0\), the stacked vector
\(( (u^\star)^\top,(w^\star)^\top )^\top\) lies in
\(\ker M^\star\). Hence, recovering \(v\) reduces to finding
\[
\begin{pmatrix}u\\w\end{pmatrix}\in\ker M^\star,
\qquad
\|u\|=1,\qquad
\|w\|^2=\eta^\star.
\]
Once such a pair is found, as per \eqref{eq:split-norm-reformulation}, the maximization of \(v^\top(C_0^{1/2}u+Bw)\) over \(\|v\|=1\) is attained at
\[
v^\star
=
\frac{C_0^{1/2}u+Bw}{\|C_0^{1/2}u+Bw\|}
=
\frac{C_0^{1/2}u+Bw}{\sqrt{T/2}},
\]
which recovers a splitting direction \(v^\star\). To confirm this, observe that
\[
\begin{aligned}
\sqrt{T/2}
&=(v^\star)^\top\bigl(C_0^{1/2}u+Bw\bigr)
\le \|C_0^{1/2}v^\star\|\,\|u\|+\|B^\top v^\star\|\,\|w\|\\
&\le \sqrt{(v^\star)^\top C_0v^\star}+\sqrt{\eta^\star}\,\|B^\top v^\star\|
\le \sqrt{T/2}.
\end{aligned}
\]
Both inequalities are therefore equalities, so \(v^\star\)
attains the directional cost \(\eta^\star\).
Using this direction in the displacement construction from
the proof of Lemma~\ref{lem:directional-split} yields an
optimal \(\U\)-structured transport plan.

\newpage
\section{Additional simulations and experiments}
\label{app:separation}

\subsection{S1--S3 SIPU datasets}

\begin{figure}[h]
\centering
\includegraphics[width=0.6\textwidth]{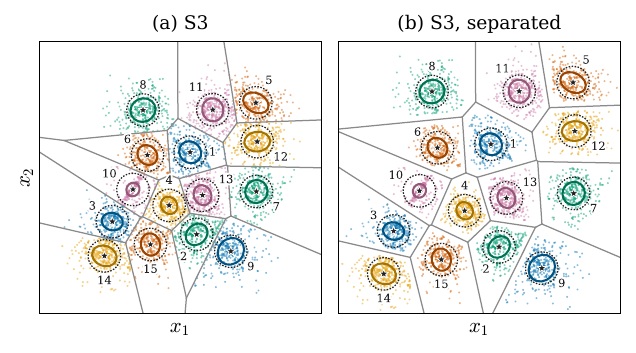}
\caption{(a) S3 and (b) S3 with cluster means scaled by \(1.25\), at \(T=0.1\) and on the same axes. The covariance ellipses and \(\sqrt{T/2}\) circles are unchanged.}
\label{fig:sipu-sep}
\end{figure}

\paragraph{S3 vs. S3 with clusters moved apart}
As a controlled check, we scale the S3 cluster means by \(1.25\), keeping each cluster's points fixed relative to its mean (Figure~\ref{fig:sipu-sep}). As expected, the smallest mass-weighted split radius, attained by split~9, remains unchanged at \(7.23\), \(8.96\), and \(8.54\) under unrestricted, horizontal, and vertical shifts (in units of \(10^{-3}\)). The smallest merge events are also unchanged (\(4{+}13\), \(4{+}13\), and \(3{+}14\)), but their mass-weighted radii increase from \(16.0\), \(16.8\), and \(19.0\) to \(36.8\), \(39.4\), and \(39.7\), roughly by a factor of \(2.1\)--\(2.3\), so, as expected, the clustering becomes more robust to merges.

\begin{figure}[t]
\centering
\includegraphics[width=0.7\textwidth]{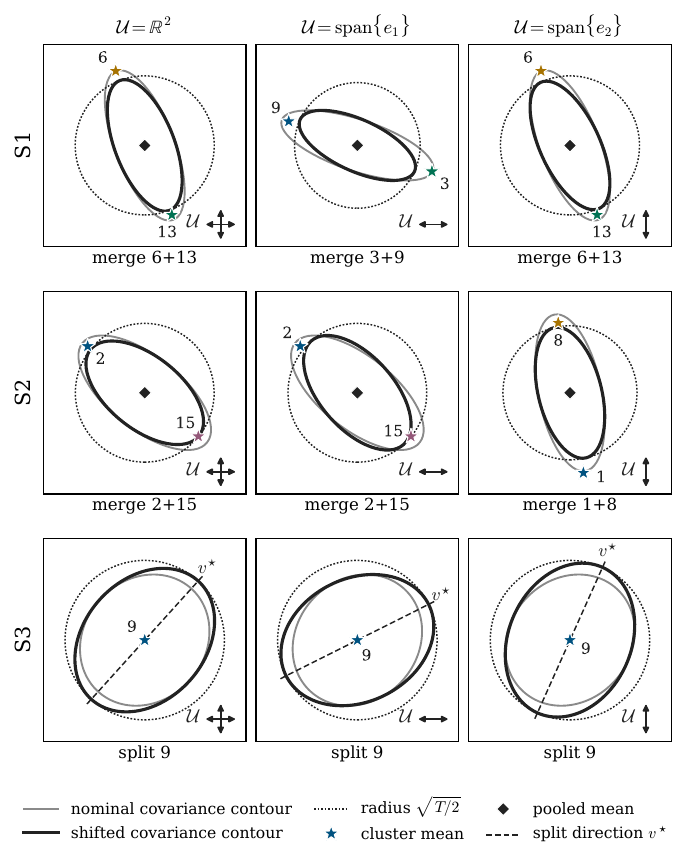}
\caption{Attaining shifts of the events realizing \(R_\U\) at \(T=0.1\) (events of Table~\ref{tab:certificates}). Each panel shows the nominal (grey) and shifted (black) covariance contours of the affected cluster or pooled pair; the shifted contours touch the \(\sqrt{T/2}\) circles.}
\label{fig:binding}
\end{figure}

\paragraph{Minimum split and merge radii}
Tables~\ref{tab:split-radii} and~\ref{tab:merge-radii} report the
minimum unweighted and mass-weighted split and merge radii.
Mass weighting can change which event is most vulnerable: under
unrestricted shifts in S1, cluster~10 has the smallest split radius,
but splitting cluster~1 requires the least population transport
distance. Figure~\ref{fig:binding} shows attaining shifts for the
certificate-attaining events.

\begin{table}[h]
\centering
\begin{tabular}{llcrcrcr}
\hline
& & \multicolumn{2}{c}{Unrestricted}
  & \multicolumn{2}{c}{Horizontal}
  & \multicolumn{2}{c}{Vertical} \\
\cline{3-8}
Dataset & Weighting
& Event & Radius & Event & Radius & Event & Radius \\
\hline
S1 & Unweighted
& Split \(10\) & 115.56
& Split \(10\) & 115.92
& Split \(10\) & 122.59 \\
& Mass-weighted
& Split \(1\)  & 29.44
& Split \(1\)  & 29.60
& Split \(10\) & 31.97 \\
\hline
S2 & Unweighted
& Split \(14\) & 75.42
& Split \(14\) & 79.78
& Split \(14\) & 86.40 \\
& Mass-weighted
& Split \(14\) & 19.52
& Split \(14\) & 20.65
& Split \(14\) & 22.36 \\
\hline
S3 & Unweighted
& Split \(9\) & 28.80
& Split \(9\) & 35.68
& Split \(9\) & 34.02 \\
& Mass-weighted
& Split \(9\) & \textbf{7.23}
& Split \(9\) & \textbf{8.96}
& Split \(9\) & \textbf{8.54} \\
\hline
\end{tabular}
\caption{Minimum split radii at \(T=0.1\).
Unweighted rows minimize \(\rho_{\mathrm{s},\U}(q_0^j;T)\)
over clusters \(j\); mass-weighted rows minimize
\(\sqrt{m_j}\rho_{\mathrm{s},\U}(q_0^j;T)\).
All values are reported in units of \(10^{-3}\).
Bold values attain the clustering certificate \(R_\U\).}
\label{tab:split-radii}
\end{table}

\begin{table}[h]
\centering
\setlength{\tabcolsep}{5pt}
\begin{tabular}{llcrcrcr}
\hline
& & \multicolumn{2}{c}{Unrestricted}
  & \multicolumn{2}{c}{Horizontal}
  & \multicolumn{2}{c}{Vertical} \\
\cline{3-8}
Dataset & Weighting
& Event & Radius & Event & Radius & Event & Radius \\
\hline
S1 & Unweighted
& Merge \(\{6,13\}\) & 32.78
& Merge \(\{3,9\}\)  & 64.61
& Merge \(\{6,13\}\) & 35.60 \\
& Mass-weighted
& Merge \(\{6,13\}\) & \textbf{12.08}
& Merge \(\{3,9\}\)  & \textbf{23.38}
& Merge \(\{6,13\}\) & \textbf{13.12} \\
& Feasible pairs
& \multicolumn{2}{c}{\(34/34\)}
& \multicolumn{2}{c}{\(13/34\)}
& \multicolumn{2}{c}{\(14/34\)} \\
\hline
S2 & Unweighted
& Merge \(\{2,15\}\) & 33.36
& Merge \(\{2,15\}\) & 45.76
& Merge \(\{1,8\}\)  & 43.34 \\
& Mass-weighted
& Merge \(\{2,15\}\) & \textbf{12.16}
& Merge \(\{2,15\}\) & \textbf{16.69}
& Merge \(\{1,8\}\)  & \textbf{15.49} \\
& Feasible pairs
& \multicolumn{2}{c}{\(33/33\)}
& \multicolumn{2}{c}{\(14/33\)}
& \multicolumn{2}{c}{\(14/33\)} \\
\hline
S3 & Unweighted
& Merge \(\{4,13\}\) & 42.20
& Merge \(\{4,13\}\) & 44.44
& Merge \(\{3,14\}\) & 52.12 \\
& Mass-weighted
& Merge \(\{4,13\}\) & 15.98
& Merge \(\{4,13\}\) & 16.83
& Merge \(\{3,14\}\) & 18.95 \\
& Feasible pairs
& \multicolumn{2}{c}{\(35/35\)}
& \multicolumn{2}{c}{\(14/35\)}
& \multicolumn{2}{c}{\(14/35\)} \\
\hline
\end{tabular}
\caption{Minimum merge radii at \(T=0.1\).
Unweighted rows minimize \(\rho_{\mathrm{m},\U}(q_0^A;T)\)
over monitored pairs \(A\); mass-weighted rows minimize
\(\sqrt{m_A}\rho_{\mathrm{m},\U}(q_0^A;T)\).
All radii are reported in units of \(10^{-3}\).
Bold values attain the clustering certificate \(R_\U\).
Feasible-pair counts give the number of finite merge radii
out of all monitored pairs.}
\label{tab:merge-radii}
\end{table}

For unrestricted shifts across S1--S3,
Table~\ref{tab:event-rankings} reports the three smallest split
and merge radii, ranked separately before and after mass weighting,
with ratios to their respective minima.

\begin{table}[!htbp]
\centering
\begin{tabular}{llccrrcrr}
\hline
& & & \multicolumn{3}{c}{Splits}
      & \multicolumn{3}{c}{Merges} \\
\cline{4-6}\cline{7-9}
Dataset & Weighting & Rank
& Event & Radius & Ratio
& Event & Radius & Ratio \\
\hline
S1 & Unweighted & 1
& Split \(10\) & 115.56 & 1.0000
& Merge \(\{6,13\}\) & 32.78 & 1.0000 \\
& & 2
& Split \(11\) & 115.59 & 1.0002
& Merge \(\{2,5\}\) & 39.13 & 1.1939 \\
& & 3
& Split \(1\) & 120.79 & 1.0453
& Merge \(\{1,8\}\) & 48.91 & 1.4921 \\
& Mass-weighted & 1
& Split \(1\) & 29.44 & 1.0000
& Merge \(\{6,13\}\) & 12.08 & 1.0000 \\
& & 2
& Split \(10\) & 30.14 & 1.0236
& Merge \(\{2,5\}\) & 14.03 & 1.1618 \\
& & 3
& Split \(11\) & 30.41 & 1.0328
& Merge \(\{1,8\}\) & 17.39 & 1.4396 \\
\hline
S2 & Unweighted & 1
& Split \(14\) & 75.42 & 1.0000
& Merge \(\{2,15\}\) & 33.36 & 1.0000 \\
& & 2
& Split \(12\) & 86.12 & 1.1419
& Merge \(\{1,8\}\) & 42.29 & 1.2677 \\
& & 3
& Split \(5\) & 91.22 & 1.2096
& Merge \(\{5,15\}\) & 43.17 & 1.2943 \\
& Mass-weighted & 1
& Split \(14\) & 19.52 & 1.0000
& Merge \(\{2,15\}\) & 12.16 & 1.0000 \\
& & 2
& Split \(12\) & 22.49 & 1.1521
& Merge \(\{1,8\}\) & 15.12 & 1.2427 \\
& & 3
& Split \(5\) & 23.51 & 1.2042
& Merge \(\{5,15\}\) & 15.86 & 1.3040 \\
\hline
S3 & Unweighted & 1
& Split \(9\) & 28.80 & 1.0000
& Merge \(\{4,13\}\) & 42.20 & 1.0000 \\
& & 2
& Split \(5\) & 38.87 & 1.3493
& Merge \(\{3,14\}\) & 50.54 & 1.1977 \\
& & 3
& Split \(12\) & 45.58 & 1.5825
& Merge \(\{6,10\}\) & 53.78 & 1.2746 \\
& Mass-weighted & 1
& Split \(9\) & 7.23 & 1.0000
& Merge \(\{4,13\}\) & 15.98 & 1.0000 \\
& & 2
& Split \(5\) & 9.39 & 1.2992
& Merge \(\{3,14\}\) & 18.37 & 1.1499 \\
& & 3
& Split \(12\) & 11.69 & 1.6173
& Merge \(\{6,10\}\) & 19.89 & 1.2449 \\
\hline
\end{tabular}
\caption{Three smallest split and merge radii under unrestricted
shifts at \(T=0.1\). Rankings are computed separately for each
dataset, event type, and weighting. Radii are reported in units
of \(10^{-3}\); mass-weighted radii are population transport
distances \(\mathrm{W}_\U\). Each ratio is relative to the minimum in its own
ranking. Merge rankings include
only monitored neighboring pairs. Event labels match
Figure~\ref{fig:sipu}.}
\label{tab:event-rankings}
\end{table}

\subsection{Higher-dimensional example: DIM64}

We consider the DIM64 benchmark from the same SIPU repository, with \(1024\) points in \(\R^{64}\) and \(16\) clusters. Centering, normalization, and the Lloyd initialization are the same as for S1--S3. The resulting labels are held fixed, and we use the same temperature \(T=0.1\). Feasible radius problems are solved using SCS through CVXPY. If a merge fails the feasibility condition~\eqref{eq:merge-feasibility},
no SDP solve is required.

\begin{table}[!htbp]
\centering
\begin{tabular}{crrrlc}
\hline
\(\ell\)
& Weighted split
& Weighted merge
& \(R_{\U_\ell}\)
& Attaining event
& Feasible merges \\
\hline
10 & 0.053610 & \(\infty\) & 0.053610
   & Split \(13\) & 0/120 \\
30 & 0.052094 & \(\infty\) & 0.052094
   & Split \(13\) & 0/120 \\
50 & 0.051143 & \(\infty\) & 0.051143
   & Split \(13\) & 0/120 \\
55 & 0.050981 & 0.168348 & 0.050981
   & Split \(13\) & 20/120 \\
60 & 0.050951 & 0.145684 & 0.050951
   & Split \(13\) & 96/120 \\
64 & 0.050749 & 0.132986 & 0.050749
   & Split \(13\) & 120/120 \\
\hline
\end{tabular}
\caption{DIM64 at \(T=0.1\): minimum mass-weighted split and merge
radii, robustness certificates, and attaining events.
All values are \(\mathrm{W}_\U\) distances in normalized coordinates;
\(\infty\) indicates that no monitored merge is feasible.}
\label{tab:dim64}
\end{table}

\end{document}